\documentclass{article}

\PassOptionsToPackage{numbers, compress}{natbib}

\usepackage[table]{xcolor}
\usepackage[preprint]{neurips_2026}
\usepackage{soul}

\usepackage[utf8]{inputenc} 
\usepackage[T1]{fontenc}    
\usepackage{hyperref}       
\usepackage{url}            
\usepackage{booktabs}       
\usepackage{amsfonts}       
\usepackage{nicefrac}       
\usepackage{microtype}      
\usepackage{xcolor}         
\usepackage{tikz}
\usepackage{xcolor}
\usepackage{comment}
\usepackage{amsmath}
\usepackage{xspace}
\usepackage{multirow}
\usepackage{float}
\usepackage{tabularx}
\usepackage{graphicx}
\usepackage{pifont}
\usepackage{longtable}

\usepackage{caption}
\usepackage{wrapfig}
\usepackage{enumitem}
\definecolor{oncoBlue}{HTML}{2F5FAE}
\usepackage{fontawesome5}

\newcommand{\huggingface}{\raisebox{-1.5pt}{\includegraphics[height=1.05em]{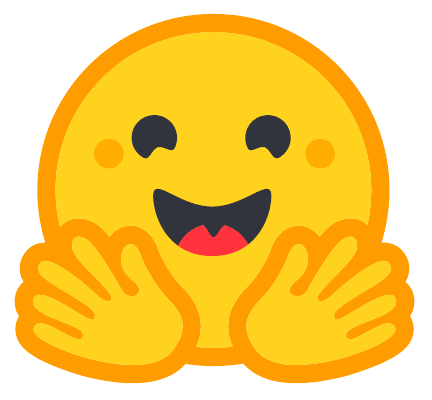}}\xspace}
\newcommand{\github}{\raisebox{-1.5pt}{\includegraphics[height=1.05em]{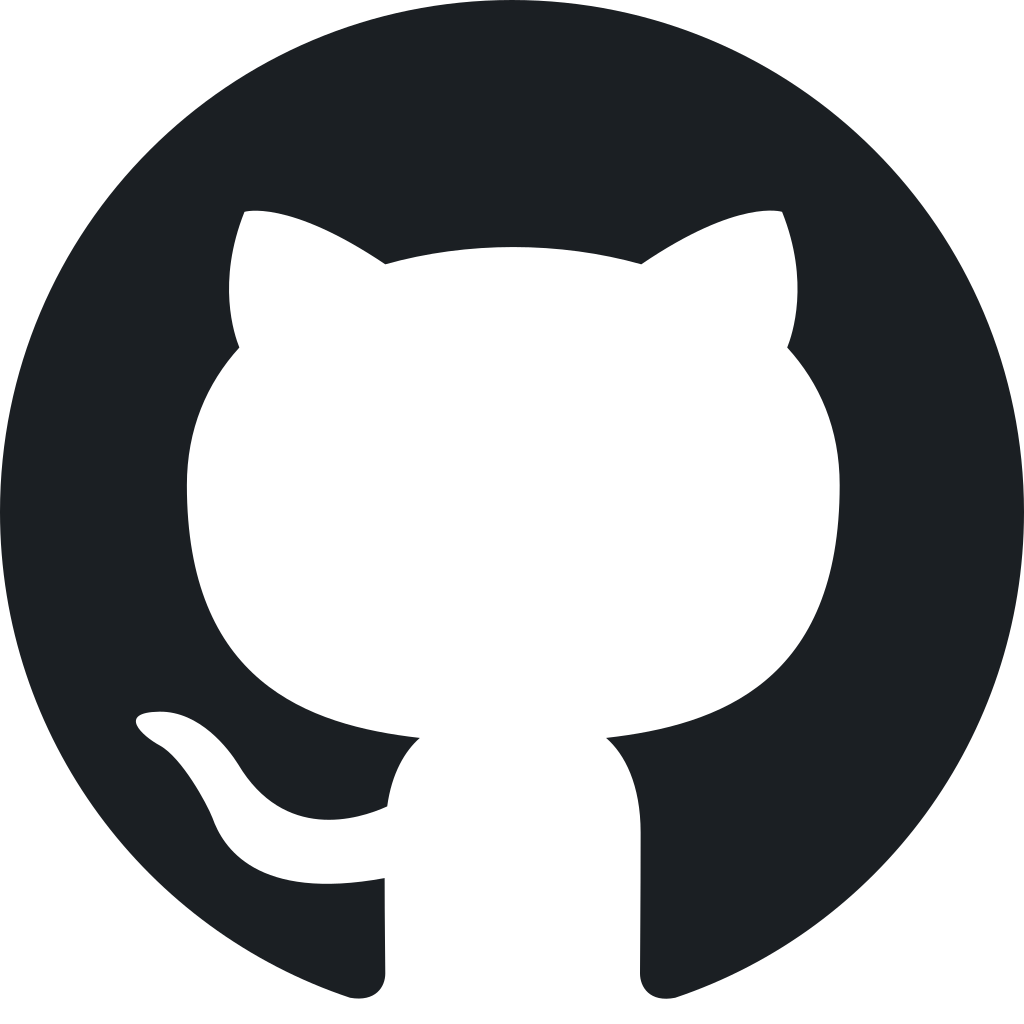}}\xspace}

\newcommand{\figcite}[1]{%
  {\sffamily\fontsize{5.7}{6.2}\selectfont\cite{#1}}%
}

\usepackage{xcolor}
\usepackage{hyperref}

\hypersetup{
    colorlinks=false,
    linkbordercolor=[HTML]{f26d6d},   
    citebordercolor=[HTML]{60CAA9},   
    urlbordercolor=[HTML]{6b9cff}     
}

\newcommand{\OncoTriadTeaserWithCites}{%
\begin{tikzpicture}
  \node[anchor=south west,inner sep=0pt] (fig) at (0,0)
    {\includegraphics[width=\linewidth,height=0.4\textheight,keepaspectratio]{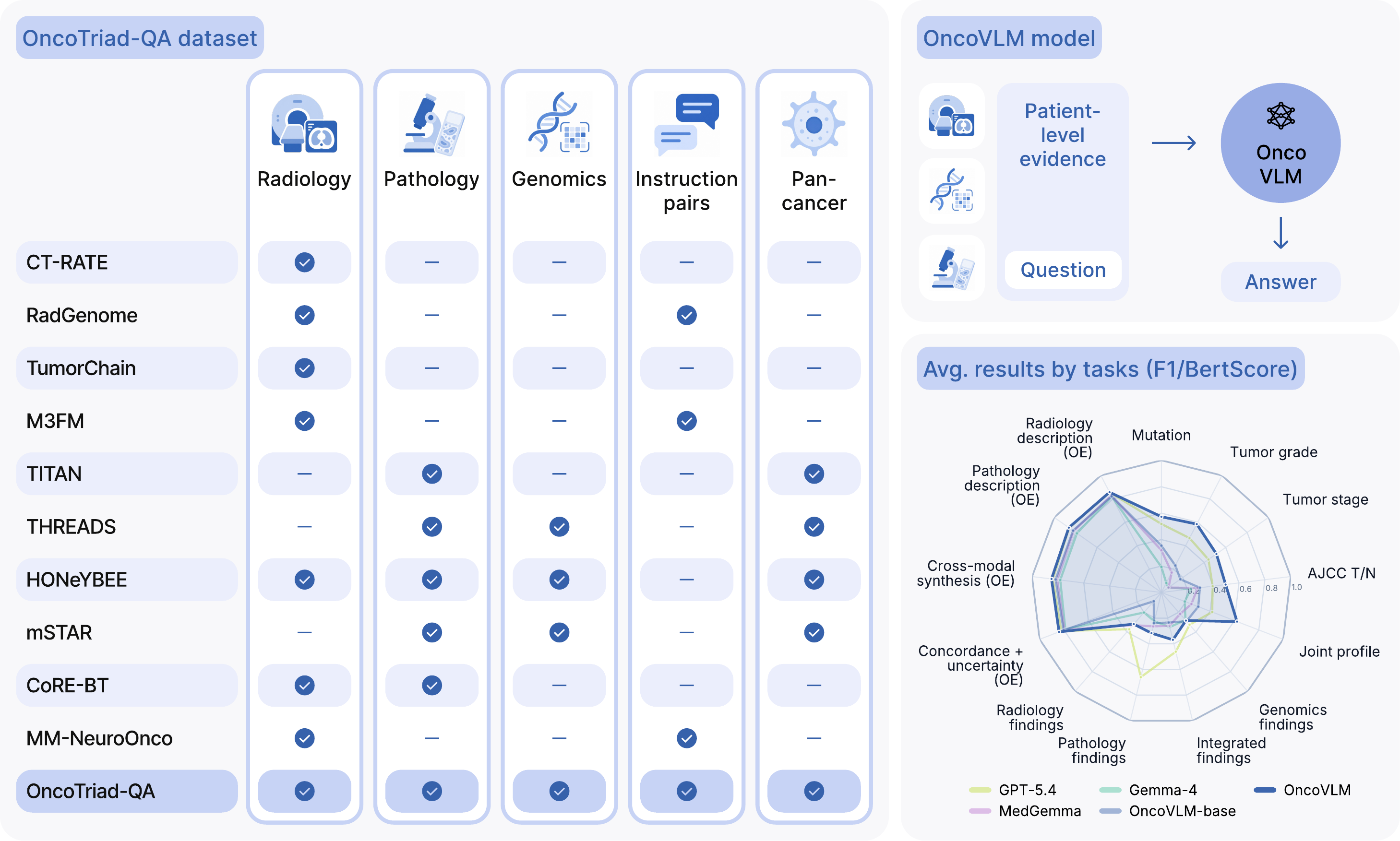}};
  \begin{scope}[x={(fig.south east)},y={(fig.north west)}]
    \node[anchor=west,inner sep=0pt] at (0.0832,0.6885) {\figcite{ct_rate}};
    \node[anchor=west,inner sep=0pt] at (0.1046,0.6254) {\figcite{radgenome}};
    \node[anchor=west,inner sep=0pt] at (0.1032,0.5627) {\figcite{tumorchain}};
    \node[anchor=west,inner sep=0pt] at (0.0661,0.5000) {\figcite{m3fm}};
    \node[anchor=west,inner sep=0pt] at (0.0643,0.4369) {\figcite{titan}};
    \node[anchor=west,inner sep=0pt] at (0.0893,0.3739) {\figcite{threads}};
    \node[anchor=west,inner sep=0pt] at (0.0971,0.3110) {\figcite{honeybee}};
    \node[anchor=west,inner sep=0pt] at (0.0721,0.2479) {\figcite{mstar}};
    \node[anchor=west,inner sep=0pt] at (0.0843,0.1852) {\figcite{corebt}};
    \node[anchor=west,inner sep=0pt] at (0.1296,0.1225) {\figcite{mm_neuro_onco}};
    \node[anchor=west,inner sep=0pt] at (0.1175,0.0595) {};
  \end{scope}
\end{tikzpicture}%
}

\title{OncoTriad-QA: A Patient-Level Radiology–Pathology–Genomics Benchmark for Pan-Cancer Reasoning}

\author{%
  Ahnaf Munir\thanks{Equal contribution.}\,\,\textsuperscript{\rm 1} \quad
  Dannong Wang\footnotemark[1]\,\,\textsuperscript{\rm 1} \quad
  Michael W. McDonald\textsuperscript{\rm 2} \\
  \textbf{Mubarak Shah}\textsuperscript{\rm 1} \quad
  \textbf{Pegah Khosravi}\textsuperscript{\rm 1,3} \quad
  \textbf{Yu Tian}\thanks{Corresponding author. Senior authorship shared with Pegah Khosravi and Mubarak Shah.}\,\,\textsuperscript{\rm 1} \\
  \\
  \textsuperscript{\rm 1}Institute of Artificial Intelligence, University of Central Florida \\
  \textsuperscript{\rm 2}AdventHealth Medical Group Urology at Celebration, FL \\
  \textsuperscript{\rm 3}Department of Clinical Sciences, College of Medicine, University of Central Florida \\
  \texttt{\{ahnaf.munir, da304044, shah, pegah.khosravi, yu.tian2\}@ucf.edu} \\
  \texttt{michael.mcdonald.md@adventhealth.com}
}

\begin{document}

\maketitle

\begin{abstract} 
Cancer diagnosis and characterization require integrating complementary evidence from radiology, pathology, genomics, and clinical metadata. However, most medical large language model (LLM) and vision-language model (VLM) benchmarks focus on isolated modalities or narrow image-text tasks, leaving patient-level oncology assessment across multiple evidence streams largely untested. 
We introduce OncoTriad-QA, a patient-level radiology–pathology–genomics benchmark for pan-cancer question answering. 
OncoTriad-QA contains 86.1k semantic questions across 9,281 TCGA patient cases from 32 cancer cohorts, aligning CT/MRI radiology, whole-slide histopathology, somatic mutations, copy-number alterations, DNA methylation, bulk RNA-seq, and clinical metadata. 
Case-specific annotations are constructed through a source-grounded LLM-assisted pipeline using curated labels, diagnostic reports, molecular profiles, and modality-derived evidence as primary sources of truth, with automated consistency checks and clinician review.
We also introduce OncoVLM, a reference multimodal model that maps modality-native radiology, pathology, DNA methylation, and RNA-seq evidence into an LLM interface through learned projectors.
Experiments show that existing general-purpose and medical LLMs remain limited on comprehensive pan-cancer QA, especially when questions require integrating imaging findings, tumor morphology, and molecular evidence. 
After fine-tuning on OncoTriad-QA, OncoVLM exceeds MedGemma-4B by an average of 10.7 points when using MCQ accuracy and BERTScore-F1, with consistent gains across multiple-choice and open-ended questions under radiology-only, pathology-only, and all-available settings.
These results demonstrate the benchmark’s value for training and evaluating models for integrated cancer question answering. 
\end{abstract}

 \begin{center}
     \vspace{-1.2em}
     \github \textbf{Code}: ~\texttt{\href{https://github.com/AI-MIND-Lab/OncoTriad-QA}{OncoTriad-QA}} \hspace{.5em}
     \huggingface \textbf{Dataset}:~\texttt{\href{https://huggingface.co/datasets/ai-mind-lab/OncoTriad-QA}{ai-mind-lab/OncoTriad-QA}}
 \end{center}

\section{Introduction}
\label{sec:intro}
Cancer exhibits substantial heterogeneity across organ systems, histologic subtypes, and molecular states, causing patients with superficially similar tumors to follow markedly different clinical trajectories \citep{marusyk2012, gerlinger2012}. Clinical cancer characterization therefore relies on evidence across multiple scales: macroscopic anatomy from computed tomography (CT) / magnetic resonance imaging (MRI), microscopic architecture from whole-slide histopathology, and molecular alterations from genomic profiling \citep{boehm2021, lipkova2022}. These streams are complementary rather than interchangeable: radiology captures lesion extent and anatomic context, histopathology captures cellular morphology and tissue organization, and molecular profiling captures subtype-defining alterations and treatment relevance. A benchmark limited to one stream can measure modality-specific recognition, but not whether a model can integrate heterogeneous evidence into a unified cancer interpretation.

\begin{figure}[t]
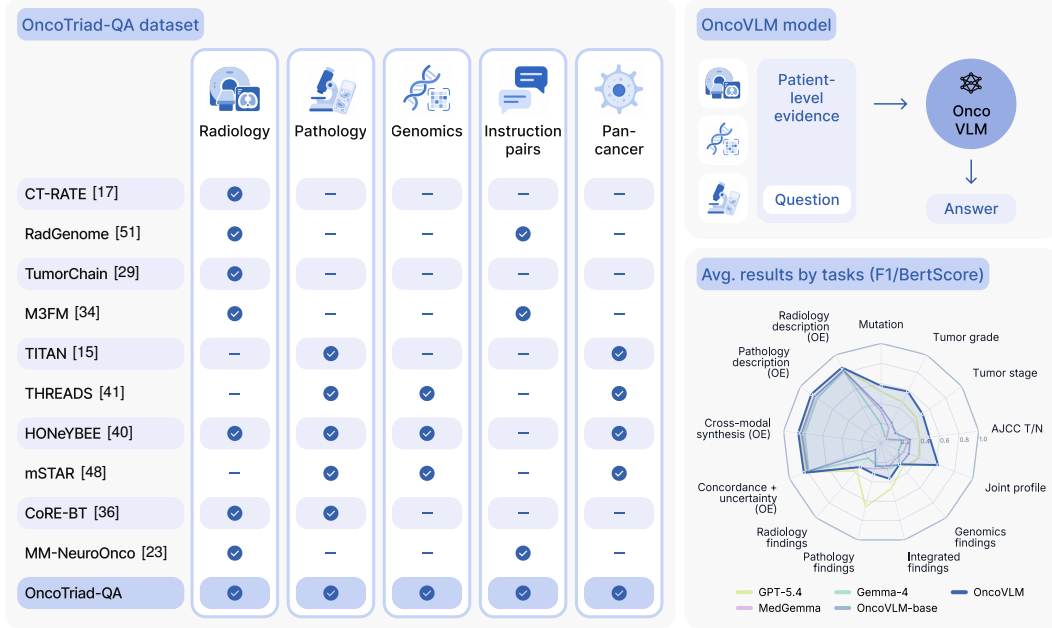

    \centering
    \OncoTriadTeaserWithCites
    \caption{Existing oncology datasets and multimodal medical resources typically cover only subsets of the evidence used in cancer characterization. OncoTriad-QA supports patient-level integration of radiology, pathology, genomics, clinical metadata, and instruction-style QA across pan-cancer cohorts, enabling evaluation of modality-specific, cross-modal, and missing-modality performance.}
    \label{fig:teaser_figure}
\end{figure}

Despite this progress, most medical large language model (LLM) and vision-language model (VLM) benchmarks still evaluate isolated or narrow modality combinations. Recent medical VLMs, including Med-Flamingo~\citep{medflamingo}, LLaVA-Med~\citep{llavamed}, and MedGemma~\citep{medgemma}, and modality-specific foundation models such as RadFM~\citep{radfm}, TITAN~\citep{titan}, and MUSK~\citep{musk}, have shown strong transfer within radiology, pathology, or biomedical image-text tasks. However, performance on pathology reports or captions, radiology, or single-modality image understanding does not ensure patient-level oncology understanding: a model may recognize findings in isolation but still fail to determine whether imaging appearance, tumor morphology, molecular alterations, and clinical context are concordant within the same patient. As illustrated in Figure~\ref{fig:teaser_figure}, existing oncology resources usually cover only partial evidence streams: radiology datasets often lack pathology and molecular data, pathology--genomics resources often lack radiology, and general medical instruction datasets rarely provide aligned patient-level evidence across all three modalities plus clinical metadata. This creates an evaluation gap between modality-specific recognition and the integrative interpretation required in oncology, motivating a unified case-level benchmark for modality-specific, cross-modal, and missing-modality QA.

We introduce \textbf{OncoTriad-QA}, a patient-level radiology, pathology, and genomics benchmark for pan-cancer analysis. Across approximately 9,000 TCGA patient cases from 32 cancer cohorts, OncoTriad-QA links CT/MRI radiology from TCIA~\citep{tcia}, whole-slide histopathology from TCGA \citep{tcga}, and GDC \citep{gdc} clinical and molecular profiles, including somatic mutations, copy-number alterations, DNA methylation, and bulk RNA-seq. The benchmark contains 86.1k semantic questions before modality-condition expansion, which are then materialized under available-evidence settings to support all-available, radiology-only, and pathology-only evaluation. Rather than treating each evidence stream as a separate task, OncoTriad-QA frames cancer characterization as patient-level QA, with MCQs and open-ended questions probing modality-specific recognition, molecular grounding, subtype reasoning, staging and grading, cross-modal concordance, and missing-modality robustness.

Because OncoTriad-QA uses LLM-assisted summarization and QA construction at scale, we make reliability and auditability central to the benchmark design. Each case summary and QA item is generated from structured modality-specific evidence, followed by automatic format and consistency checks and medical expert review on a subset of cases. Together with fixed patient-level splits and evaluation protocols stratified by cancer cohort and modality availability, these checks reduce unsupported claims and help assess whether models use the available patient evidence rather than single-modality shortcuts.

To demonstrate how the benchmark supports both training and evaluation, we further introduce \textbf{OncoVLM}, a reference multimodal language model trained on OncoTriad-QA. OncoVLM connects frozen radiology, pathology, methylation, and transcriptomic encoders to an LLM backbone through learned projectors and instruction tuning, allowing heterogeneous patient evidence to be represented in a unified language-model interface. We benchmark OncoVLM against general medical VLMs, and open- and closed-source multimodal LLM baselines. 
Results show that existing models remain limited on pan-cancer QA, while fine-tuned OncoVLM exceeds MedGemma-4B by approximately 10.7 points when using MCQ accuracy and BERTScore-F1 across radiology-only, pathology-only, and all-available settings.

The main contributions of our work are:
\begin{itemize}
    \item \textbf{OncoTriad-QA, a clinician-guided patient-level benchmark for multimodal pan-cancer assessment.}
    We construct a multimodal oncology benchmark that aligns imaging, histologic, molecular, and clinical evidence across approximately 9,000 patient cases from 32 cancer cohorts, with medical-expert guided prompt/question design, and case-summary audit for clinical validity. The benchmark includes MCQs and open-ended questions targeting staging and grading, modality-specific recognition, molecular grounding, and cross-modal concordance.

    \item \textbf{OncoVLM, a reference model for multimodal oncology QA.}
    Unlike many existing multimodal medical VLMs restricted to low-resolution PNG proxies, OncoVLM ingests modality-native clinical data, such as gigapixel whole-slide pathology, multi-slice radiology, and genomic sequences, through modality-specific encoders and learned projectors, and can work with multiple modality combinations and handle missing modalities.  

    \item \textbf{A universal evaluation framework of multimodal oncology.} 
    Designed to mimic real clinical settings, our protocol evaluates multimodal cancer QA across arbitrary modality combinations. It remains agnostic to VLM input design by providing fallbacks whenever a model cannot consume original full-scale imaging or molecular profiles. Using this protocol, we evaluate OncoVLM with state-of-the-art proprietary multimodal and open-weight baselines and medical VLMs.
\end{itemize}

\section{Related work}

\textbf{Medical vision-language models.} General-purpose medical VLMs have achieved strong transfer across clinical imaging tasks, with systems ranging from contrastive
pretraining on biomedical figure-caption pairs~\citep{biomedclip} to few-shot multimodal learning~\citep{medflamingo} and GPT-supervised instruction tuning~\citep{llavamed, pubmedvision, medgemma}. Despite this progress, these systems are trained on single modalities or narrow modality pairs and cannot jointly reason over radiology,
histopathology, and molecular data within a unified framework.

\textbf{Modality-specific foundation models.}
Within individual modalities, large-scale pretraining has produced capable specialists. In radiology, RadFM~\citep{radfm}, CheXagent~\citep{chexagent}, and BioViL-T~\citep{biovilt} achieve strong performance on report generation and image interpretation. In pathology, UNI~\citep{uni}, CONCH~\citep{conch}, MUSK~\citep{musk}, TITAN~\citep{titan}, GigaPath~\citep{xu2024gigapath}, and
THREADS~\citep{threads} encode rich slide-level representations, with recent models incorporating molecular supervision. In genomics, Geneformer \citep{geneformer}, scGPT~\citep{scgpt}, and BulkFormer~\citep{bulkformer} capture transcriptomic and epigenomic structure at scale. Yet none can jointly process imaging and molecular evidence at inference time, nor support open-ended interpretation across all three modalities.

\textbf{Multimodal fusion for oncology.}
Prior work has combined histopathology and molecular data for tasks such
as survival prediction and subtype classification \citep{chen2022pancancer, chen2022pathomic}. While these methods demonstrate the value of cross-modal integration, they
produce scalar task-specific outputs rather than general-purpose language-grounded responses. Like prior medical VLMs, they too remain confined to a subset of modalities, with none jointly incorporating radiology, pathology, and genomics.

\textbf{Instruction tuning in medical AI.}
Instruction tuning has proven effective for medical VLMs, as shown by MedTrinity-25M~\citep{medtrinity} and similar efforts~\citep{titan}. The complementary framework of learning with privileged information (LUPI)~\citep{shiao2014} where auxiliary signals available only at training time guide representation learning has been applied in computational pathology via TriDeNT~\citep{trident}. Our work extends both paradigms to multimodal oncology. Here, the LLM observes full multimodal inputs and structured clinical ground truth during dataset construction, while OncoVLM learns to replicate this integrated interpretation from encoder-derived representations alone at inference time.

\section{Dataset construction}
\label{sec:dataset}
OncoTriad-QA is a patient-level multimodal benchmark spanning approximately 9,000 cases from 32 cancer cohorts. For each case, we align available TCIA radiology, TCGA-derived whole-slide histopathology, and GDC molecular and clinical data, including somatic mutations, copy-number alterations, DNA methylation, and bulk RNA-seq. Because modality availability with each patient varies in the real world clinical setting~\citep{wang2026handlinginterpretingmissingmodalities, Zhang_2022}, we organize the benchmark into all-available, radiology-only, and pathology-only settings, supporting both complete multimodal evaluation and realistic missing-modality scenarios. GPT-5.4~\citep{gpt54} is used for information summarization and case-specific question construction. 
In total, OncoTriad-QA contains approximately 86.1k distinct semantic questions, including 43.2k universal MCQs, 21.2k case-specific MCQs, and 21.7k case-specific open-ended questions. After modality-condition expansion, these yield 182.6k materialized question rows.

\begin{figure}
    \centering
    \includegraphics[width=1\linewidth]{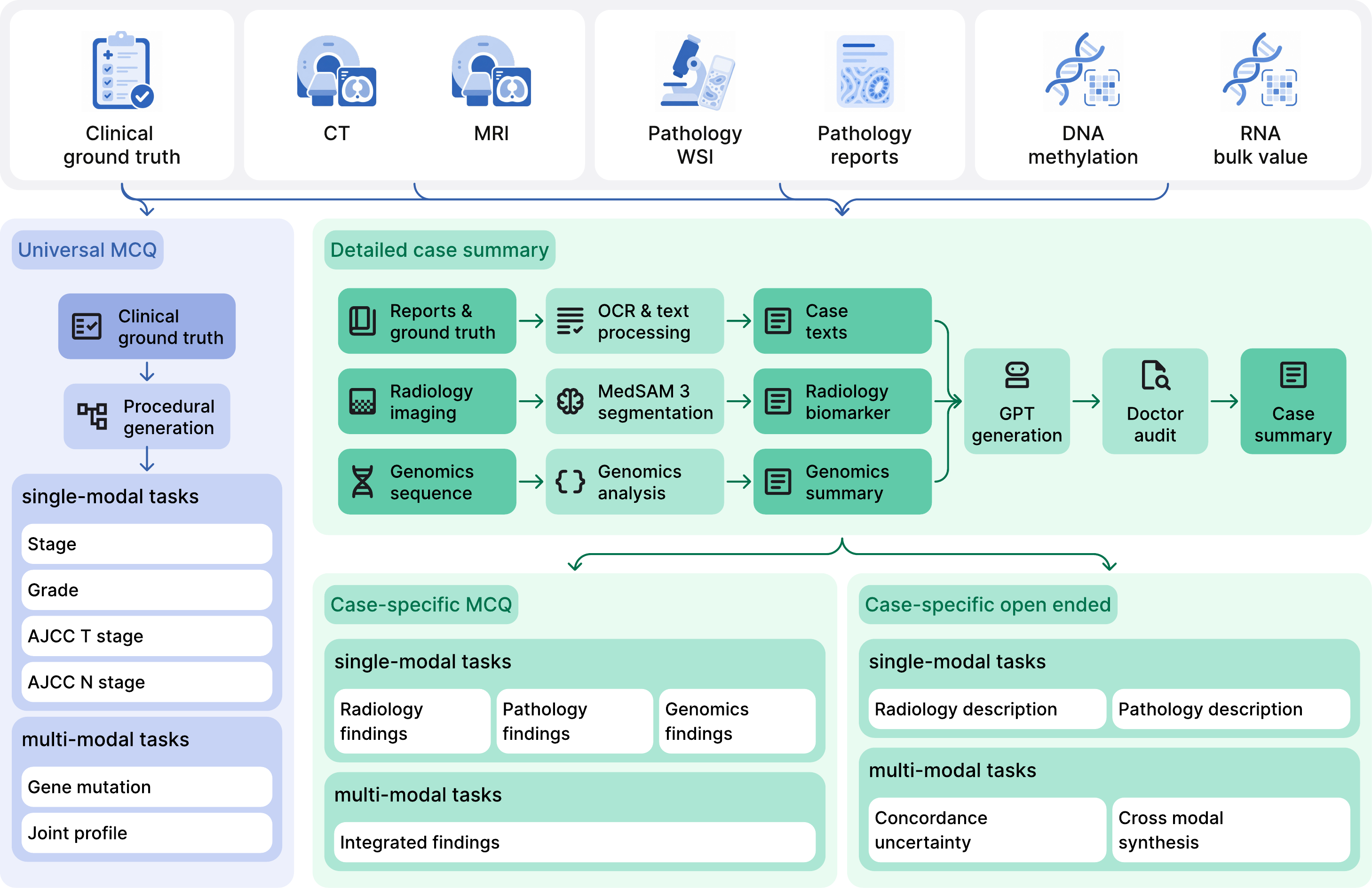}
    \caption{
    Overview of OncoTriad-QA construction. Clinical metadata, radiology, pathology, DNA methylation, and bulk RNA-seq are converted into structured patient-level evidence. Clinical variables generate universal MCQs, while modality-specific processing supports GPT-based case summaries and QA generation. The final benchmark includes case-specific MCQs and open-ended QAs across radiology, pathology, genomics, integrated findings, cross-modal synthesis, and uncertainty, with physician audit for reliability.
    }
    \label{fig:dataset_construction}
\end{figure}

\textbf{Detailed case summary.}
We construct a structured case representation from diagnostic reports, clinical metadata, curated ground-truth labels, and modality-derived signals. This representation serves as the canonical context for case-specific tasks, making generated questions traceable to source evidence and supporting consistency checks and expert audit.
(i) \textbf{Radiology.}
CT and MRI tumor regions are segmented with Medical SAM3~\citep{medicalsam3} and treated as ROIs. For each case, imaging series are ranked by segmentation consistency and mean tumor area; we retain the top-three and extract representative slices per series. Geometric, intensity, and shape biomarkers are summarized at the series level and aggregated with rank-based weights, while segmentation masks are converted to bounding boxes for language-model annotation.
(ii) \textbf{Pathology.}
Pathology reports and ground-truth labels provide the primary evidence for tumor morphology, while representative ROI images from TCGA-UniformTumor-8K~\citep{titan} supplement the text with visual patterns that may not be fully captured in the reports. Because current closed-source LLMs cannot directly process gigapixel whole-slide images (WSIs), GPT is given ROI images rather than full WSIs.
(iii) \textbf{Genomics.}
Molecular data are converted into structured genomic text summarizing RNA expression, DNA methylation, somatic mutations, copy-number alterations, pathway activity, immune and stromal signatures, molecular subtype indicators, driver-gene status, tumor mutational burden, MSI status, and HRD score.

\textbf{Universal MCQs.} 
We procedurally generate universal MCQs from curated structured labels using standardized templates that can be instantiated whenever the required patient-level variable is available. These questions cover tumor stage, tumor grade, AJCC T/N category, mutation status, and joint stage-mutation profiles. Because answers are filled directly from curated ground-truth fields, this stream provides a controlled evaluation setting independent of LLM question construction.

\textbf{Case-specific MCQs and open-ended questions.} 
Case-specific QA is generated from detailed case summaries that integrate radiology, pathology, genomics, clinical labels, and cross-modal interpretation. For open-ended QA, the GPT uses these summaries to produce concise answers covering modality-specific findings, integrated diagnosis, concordance, synthesis, and uncertainty. For case-specific MCQs, we avoid prompting GPT to directly generate distractors, which can introduce language-pattern shortcuts \citep{computers15020130, balepur-etal-2025-best}. Instead, we first extract a contrastive fingerprint for each case by showing GPT the target case and three additional cases from the same cancer cohort. The target fingerprint becomes the correct option, while fingerprints from other same-cohort cases serve as wrong options. This makes the task depend on patient-specific evidence rather than cancer-type priors or generic answer style.

\textbf{Clinical expert review.} 
To further assess annotation quality, we obtained physician feedback on a subset of generated captions and MCQs. The reviewer highlighted that some cases lacked sufficient imaging context, complete stage or grade information, lymph-node status, or fully informative genomic evidence, making certain staging, prognosis, or treatment-oriented questions difficult to support. Based on this feedback, we refined the generation guidelines to emphasize evidence-grounded wording, explicit uncertainty when case information is incomplete, and avoidance of unsupported clinical conclusions.

\section{Model architecture}
\label{sec:model_arch}

\begin{figure}
  \centering
    \includegraphics[width=1\linewidth]{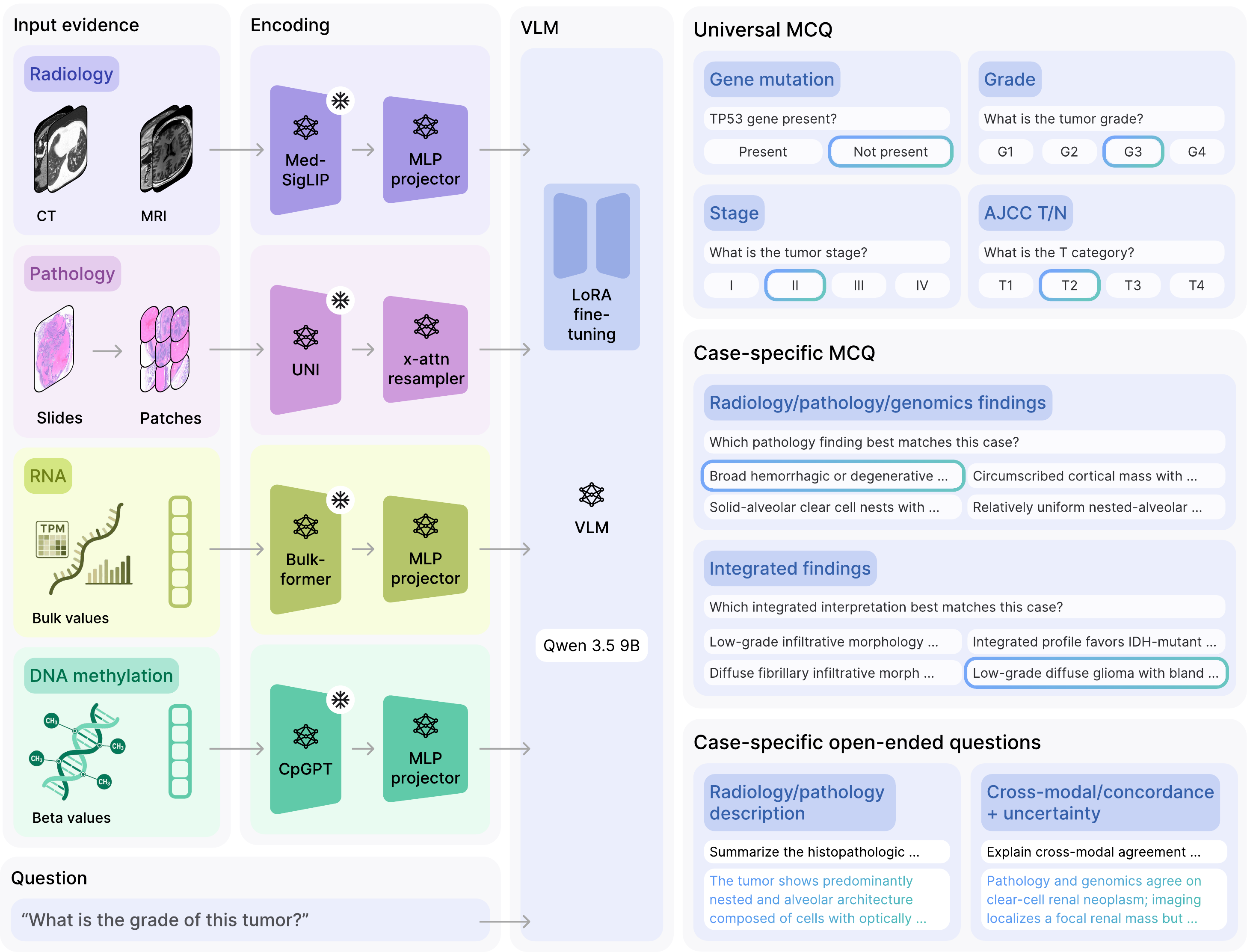}
    \caption{Overview of the proposed multimodal architecture (OncoVLM). Modality-specific encoders map radiology, pathology, and genomic inputs into a shared representation space, which is processed by a LoRA-tuned vision-language model to perform classification and multimodal tasks.}  
    \label{fig:architecture}
\end{figure}

OncoTriad-QA supports both evaluation and supervised training of patient-level multimodal oncology models. Alongside the benchmark, we introduce OncoVLM, a model designed to use modality-native oncology evidence rather than only text summaries or low-resolution image proxies. OncoVLM encodes multi-slice radiology series, gigapixel whole-slide pathology patch sets, DNA methylation beta-value profiles, and bulk RNA-seq expression profiles, then maps these heterogeneous inputs into a shared language-model interface.

\textbf{Modality encoding and projection.}
For each patient case, the available evidence is processed through frozen modality-specific encoders. Radiology inputs are encoded with MedSigLIP-448~\citep{medgemma}; Pathology WSIs are encoded with pathology foundation model UNI~\citep{uni}; RNA-seq expression and DNA methylation beta values are encoded with BulkFormer~\citep{bulkformer} and CpGPT~\citep{cpgpt}, respectively. Keeping these encoders frozen preserves modality-specific representations learned from large-scale pretraining and reduces overfitting across heterogeneous cancer cohorts \citep{llavamed}.
Each encoder is paired with a learned projector that maps its outputs into the language model embedding space. Since pathology slides produce variable-length and long sets of patch embeddings, the pathology branch uses a Perceiver-style resampler~\citep{jaegle2021perceiver, NEURIPS2022_960a172b} to compress slide-level evidence into a fixed number of tokens, rather than a simple MLP. Radiology, DNA methylation, and RNA-seq use modality-specific MLP projectors for slice-level or case-level embeddings. The projected tokens are inserted into the instruction prompt with modality tags. Missing modalities are handled by omitting the corresponding token blocks, allowing the model to operate under the same evidence configurations used in OncoTriad-QA, including all-available, pathology-only, and radiology-only settings.

\textbf{Two-stage training.}
In the modality-alignment stage, each projector is trained independently on modality-specific captions while the corresponding encoder and the language model remain frozen. This maps each frozen biomedical representation into the LLM embedding space before any joint VQA training, giving each modality a stable language interface and reducing early cross-modal interference \citep{pmlr-v162-huang22e}. In the supervised fine-tuning (SFT) stage, the encoders and aligned projectors are frozen, and the LLM is adapted with low-rank adaptation (LoRA) on OncoTriad-QA training split. We use the standard autoregressive next-token cross-entropy loss over target answer tokens, with the instruction prompt and modality tokens used as conditioning context.

\section{Experiments}
\label{sec:results}

We use Qwen3.5-9B as the OncoVLM language model backbone and evaluate the following: 
(i) \textbf{GPT-5.4}~\citep{gpt54}, a proprietary general-purpose model used as a strong reference rather than a directly comparable trainable system;
(ii) \textbf{MedGemma-4B}~\citep{medgemma}, a medical vision-language foundation model;
(iii) \textbf{Gemma-4-E4B}~\citep{gemma_4_2026}, a recent frontier open-weight model;
(iv) \textbf{OncoVLM-base}, OncoVLM with trained projectors but without the fine-tuning stage;
(v) \textbf{OncoVLM-finetuned}, the OncoVLM-base model fine-tuned on OncoTriad-QA. 
Baselines receive representative pathology ROI images, representative radiology slices, and structured molecular text summaries when available, whereas OncoVLM receives encoder-derived modality representations. 
We evaluate on the held-out OncoTriad-QA test split. MCQs use accuracy and macro-F1, while open-ended QAs use BERTScore-F1, ROUGE-L, and an LLM-as-judge evaluation on BRCA. Claude Sonnet 4.6 \citep{anthropic2025sonnet46} is used as the judge to compare OncoVLM against GPT-5.4 pairwise. We report OncoVLM win rate, counting ties as half wins. Table~\ref{tab:main_result} results are averaged over three independent runs with standard deviations reported. 

\begin{table}[t]
\captionsetup{type=table}
\captionof{table}{OncoTriad-QA benchmark results by task category on the all-available modality setting. Values are mean and standard deviation over three separate run, and using different seeds on non-GPT model.}
\label{tab:main_result}
\centering
\small
\setlength{\tabcolsep}{5pt}
\renewcommand{\arraystretch}{1.18}
\providecommand{\cellmetric}[3]{}
\renewcommand{\cellmetric}[3]{\makebox[\linewidth][c]{\begin{tabular}[t]{@{}c@{}}#1\\[-0.30em]{\scriptsize\textcolor[HTML]{#3}{#2}}\end{tabular}}}
\providecommand{\bertcell}[1]{#1}
\begin{tabular}{>{\raggedright\arraybackslash}p{\dimexpr\linewidth-1.9cm-1.9cm-1.9cm-1.9cm-1.9cm-12\tabcolsep\relax}>{\centering\arraybackslash}p{1.9cm}>{\centering\arraybackslash}p{1.9cm}>{\centering\arraybackslash}p{1.9cm}>{\centering\arraybackslash}p{1.9cm}>{\centering\arraybackslash}p{1.9cm}}
\toprule
\rowcolor[HTML]{F3F5FC}
\textbf{Name} & GPT-5.4 & \cellcolor[HTML]{EFF1FC}MedGemma & \cellcolor[HTML]{EFF1FC}Gemma-4 & \cellcolor[HTML]{EFF1FC}OncoVLM & \cellcolor[HTML]{EBEEFC}OncoVLM \\
\rowcolor[HTML]{F3F5FC}
 & -- & \cellcolor[HTML]{EFF1FC}4B & \cellcolor[HTML]{EFF1FC}E4B & \cellcolor[HTML]{EFF1FC}Qwen-3.5-9B & \cellcolor[HTML]{EBEEFC}Qwen-3.5-9B \\
\rowcolor[HTML]{F3F5FC}
\textbf{Finetuned} & \faMinus & \cellcolor[HTML]{EFF1FC}\faMinus & \cellcolor[HTML]{EFF1FC}\faMinus & \cellcolor[HTML]{EFF1FC}\faMinus & \cellcolor[HTML]{EBEEFC}\faCheckCircle \\
\rowcolor[HTML]{F3F5FC}
\textbf{Use projector} & \faMinus & \cellcolor[HTML]{EFF1FC}\faMinus & \cellcolor[HTML]{EFF1FC}\faMinus & \cellcolor[HTML]{EFF1FC}\faCheckCircle & \cellcolor[HTML]{EBEEFC}\faCheckCircle \\
\arrayrulecolor[HTML]{000000}\hline\arrayrulecolor{black}
\rowcolor[HTML]{F3F5FC}
\multicolumn{6}{l}{\rule[-0.8ex]{0pt}{3.2ex}\textbf{Universal MCQ} \textnormal{(Accuracy\% / \textcolor[HTML]{3661AB}{F1})}} \\
\arrayrulecolor[HTML]{000000}\hline\arrayrulecolor{black}
\noalign{\vskip 0.28em}
Mutation & \cellmetric{\underline{51.5} / \textcolor[HTML]{3661AB}{\underline{0.515}}}{$\pm$ 1.2 / $\pm$ 0.01}{64748B} & \cellmetric{46.0 / \textcolor[HTML]{3661AB}{0.317}}{$\pm$ 0.0 / $\pm$ 0.00}{64748B} & \cellmetric{11.5 / \textcolor[HTML]{3661AB}{0.190}}{$\pm$ 0.2 / $\pm$ 0.00}{64748B} & \cellmetric{44.2 / \textcolor[HTML]{3661AB}{0.344}}{$\pm$ 0.3 / $\pm$ 0.01}{64748B} & \cellcolor[HTML]{EBEEFC}\cellmetric{\textbf{58.2} / \textcolor[HTML]{3661AB}{\textbf{0.577}}}{$\pm$ 1.3 / $\pm$ 0.01}{64748B} \\
Tumor grade & \cellmetric{\underline{52.2} / \textcolor[HTML]{3661AB}{\underline{0.384}}}{$\pm$ 2.7 / $\pm$ 0.03}{64748B} & \cellmetric{40.3 / \textcolor[HTML]{3661AB}{0.162}}{$\pm$ 0.5 / $\pm$ 0.02}{64748B} & \cellmetric{10.3 / \textcolor[HTML]{3661AB}{0.073}}{$\pm$ 1.3 / $\pm$ 0.01}{64748B} & \cellmetric{37.5 / \textcolor[HTML]{3661AB}{0.178}}{$\pm$ 0.8 / $\pm$ 0.00}{64748B} & \cellcolor[HTML]{EBEEFC}\cellmetric{\textbf{64.7} / \textcolor[HTML]{3661AB}{\textbf{0.615}}}{$\pm$ 0.5 / $\pm$ 0.05}{64748B} \\
Tumor stage & \cellmetric{\underline{44.5} / \textcolor[HTML]{3661AB}{\underline{0.464}}}{$\pm$ 1.0 / $\pm$ 0.01}{64748B} & \cellmetric{7.7 / \textcolor[HTML]{3661AB}{0.074}}{$\pm$ 0.3 / $\pm$ 0.00}{64748B} & \cellmetric{5.7 / \textcolor[HTML]{3661AB}{0.053}}{$\pm$ 0.8 / $\pm$ 0.01}{64748B} & \cellmetric{18.8 / \textcolor[HTML]{3661AB}{0.180}}{$\pm$ 1.3 / $\pm$ 0.02}{64748B} & \cellcolor[HTML]{EBEEFC}\cellmetric{\textbf{51.8} / \textcolor[HTML]{3661AB}{\textbf{0.480}}}{$\pm$ 2.5 / $\pm$ 0.03}{64748B} \\
AJCC T/N & \cellmetric{\underline{42.3} / \textcolor[HTML]{3661AB}{\underline{0.400}}}{$\pm$ 1.5 / $\pm$ 0.02}{64748B} & \cellmetric{36.4 / \textcolor[HTML]{3661AB}{0.309}}{$\pm$ 0.2 / $\pm$ 0.00}{64748B} & \cellmetric{39.2 / \textcolor[HTML]{3661AB}{0.213}}{$\pm$ 0.1 / $\pm$ 0.00}{64748B} & \cellmetric{39.6 / \textcolor[HTML]{3661AB}{0.310}}{$\pm$ 1.4 / $\pm$ 0.01}{64748B} & \cellcolor[HTML]{EBEEFC}\cellmetric{\textbf{60.0} / \textcolor[HTML]{3661AB}{\textbf{0.520}}}{$\pm$ 1.1 / $\pm$ 0.00}{64748B} \\
Joint profile & \cellmetric{\underline{38.9} / \textcolor[HTML]{3661AB}{\underline{0.388}}}{$\pm$ 2.3 / $\pm$ 0.02}{64748B} & \cellmetric{28.9 / \textcolor[HTML]{3661AB}{0.246}}{$\pm$ 0.3 / $\pm$ 0.00}{64748B} & \cellmetric{18.2 / \textcolor[HTML]{3661AB}{0.197}}{$\pm$ 1.0 / $\pm$ 0.01}{64748B} & \cellmetric{31.9 / \textcolor[HTML]{3661AB}{0.314}}{$\pm$ 0.6 / $\pm$ 0.01}{64748B} & \cellcolor[HTML]{EBEEFC}\cellmetric{\textbf{62.4} / \textcolor[HTML]{3661AB}{\textbf{0.623}}}{$\pm$ 1.2 / $\pm$ 0.01}{64748B} \\
\textbf{Mean} & \cellmetric{\underline{46.0} / \textcolor[HTML]{3661AB}{\underline{0.449}}}{$\pm$ 1.6 / $\pm$ 0.02}{64748B} & \cellmetric{36.0 / \textcolor[HTML]{3661AB}{0.267}}{$\pm$ 0.2 / $\pm$ 0.00}{64748B} & \cellmetric{17.9 / \textcolor[HTML]{3661AB}{0.177}}{$\pm$ 0.5 / $\pm$ 0.01}{64748B} & \cellmetric{37.6 / \textcolor[HTML]{3661AB}{0.306}}{$\pm$ 0.7 / $\pm$ 0.01}{64748B} & \cellcolor[HTML]{EBEEFC}\cellmetric{\textbf{59.4} / \textcolor[HTML]{3661AB}{\textbf{0.571}}}{$\pm$ 1.3 / $\pm$ 0.01}{64748B} \\
\addlinespace[0.35em]
\arrayrulecolor[HTML]{000000}\hline\arrayrulecolor{black}
\rowcolor[HTML]{F3F5FC}
\multicolumn{6}{l}{\rule[-0.8ex]{0pt}{3.2ex}\textbf{Case-specific MCQ} \textnormal{(Accuracy\% / \textcolor[HTML]{3661AB}{F1})}} \\
\arrayrulecolor[HTML]{000000}\hline\arrayrulecolor{black}
\noalign{\vskip 0.28em}
Radiology findings & \cellmetric{\textbf{42.3} / \textcolor[HTML]{3661AB}{\textbf{0.419}}}{$\pm$ 8.1 / $\pm$ 0.08}{64748B} & \cellmetric{\underline{34.9} / \textcolor[HTML]{3661AB}{\underline{0.332}}}{$\pm$ 3.2 / $\pm$ 0.03}{64748B} & \cellmetric{23.8 / \textcolor[HTML]{3661AB}{0.202}}{$\pm$ 1.6 / $\pm$ 0.01}{64748B} & \cellmetric{5.8 / \textcolor[HTML]{3661AB}{0.082}}{$\pm$ 0.9 / $\pm$ 0.01}{64748B} & \cellcolor[HTML]{EBEEFC}\cellmetric{31.7 / \textcolor[HTML]{3661AB}{0.320}}{$\pm$ 4.2 / $\pm$ 0.04}{64748B} \\
Pathology findings & \cellmetric{\textbf{63.2} / \textcolor[HTML]{3661AB}{\textbf{0.630}}}{$\pm$ 1.3 / $\pm$ 0.01}{64748B} & \cellmetric{24.3 / \textcolor[HTML]{3661AB}{0.224}}{$\pm$ 0.1 / $\pm$ 0.00}{64748B} & \cellmetric{24.0 / \textcolor[HTML]{3661AB}{0.235}}{$\pm$ 1.0 / $\pm$ 0.01}{64748B} & \cellmetric{24.2 / \textcolor[HTML]{3661AB}{0.222}}{$\pm$ 0.9 / $\pm$ 0.01}{64748B} & \cellcolor[HTML]{EBEEFC}\cellmetric{\underline{32.1} / \textcolor[HTML]{3661AB}{\underline{0.319}}}{$\pm$ 1.0 / $\pm$ 0.01}{64748B} \\
Genomics findings & \cellmetric{\textbf{30.5} / \textcolor[HTML]{3661AB}{\textbf{0.305}}}{$\pm$ 1.1 / $\pm$ 0.01}{64748B} & \cellmetric{23.6 / \textcolor[HTML]{3661AB}{0.231}}{$\pm$ 0.5 / $\pm$ 0.00}{64748B} & \cellmetric{27.4 / \textcolor[HTML]{3661AB}{0.263}}{$\pm$ 0.4 / $\pm$ 0.01}{64748B} & \cellmetric{26.1 / \textcolor[HTML]{3661AB}{0.268}}{$\pm$ 0.6 / $\pm$ 0.01}{64748B} & \cellcolor[HTML]{EBEEFC}\cellmetric{\underline{29.4} / \textcolor[HTML]{3661AB}{\underline{0.293}}}{$\pm$ 2.5 / $\pm$ 0.02}{64748B} \\
Integrated findings & \cellmetric{\textbf{47.8} / \textcolor[HTML]{3661AB}{\textbf{0.478}}}{$\pm$ 0.2 / $\pm$ 0.00}{64748B} & \cellmetric{25.7 / \textcolor[HTML]{3661AB}{0.260}}{$\pm$ 0.2 / $\pm$ 0.00}{64748B} & \cellmetric{27.9 / \textcolor[HTML]{3661AB}{0.272}}{$\pm$ 0.4 / $\pm$ 0.00}{64748B} & \cellmetric{24.4 / \textcolor[HTML]{3661AB}{0.242}}{$\pm$ 0.8 / $\pm$ 0.01}{64748B} & \cellcolor[HTML]{EBEEFC}\cellmetric{\underline{36.0} / \textcolor[HTML]{3661AB}{\underline{0.359}}}{$\pm$ 2.0 / $\pm$ 0.02}{64748B} \\
\textbf{Mean} & \cellmetric{\textbf{47.0} / \textcolor[HTML]{3661AB}{\textbf{0.470}}}{$\pm$ 1.2 / $\pm$ 0.01}{64748B} & \cellmetric{25.0 / \textcolor[HTML]{3661AB}{0.243}}{$\pm$ 0.4 / $\pm$ 0.00}{64748B} & \cellmetric{26.3 / \textcolor[HTML]{3661AB}{0.254}}{$\pm$ 0.6 / $\pm$ 0.01}{64748B} & \cellmetric{24.0 / \textcolor[HTML]{3661AB}{0.236}}{$\pm$ 0.8 / $\pm$ 0.01}{64748B} & \cellcolor[HTML]{EBEEFC}\cellmetric{\underline{32.5} / \textcolor[HTML]{3661AB}{\underline{0.323}}}{$\pm$ 1.9 / $\pm$ 0.02}{64748B} \\
\addlinespace[0.35em]
\arrayrulecolor[HTML]{000000}\hline\arrayrulecolor{black}
\rowcolor[HTML]{F3F5FC}
\multicolumn{6}{l}{\rule[-0.8ex]{0pt}{3.2ex}\textbf{Case specific open-ended} \textnormal{(BERT-F1 / \textcolor[HTML]{3661AB}{ROUGE-L})}} \\
\arrayrulecolor[HTML]{000000}\hline\arrayrulecolor{black}
\noalign{\vskip 0.28em}
Concordance + uncertainty & \cellmetric{\underline{0.823} / \textcolor[HTML]{3661AB}{\underline{0.150}}}{$\pm$ 0.00 / $\pm$ 0.00}{64748B} & \cellmetric{0.795 / \textcolor[HTML]{3661AB}{0.120}}{$\pm$ 0.00 / $\pm$ 0.00}{64748B} & \cellmetric{0.788 / \textcolor[HTML]{3661AB}{0.103}}{$\pm$ 0.00 / $\pm$ 0.00}{64748B} & \cellmetric{0.802 / \textcolor[HTML]{3661AB}{0.127}}{$\pm$ 0.00 / $\pm$ 0.00}{64748B} & \cellcolor[HTML]{EBEEFC}\cellmetric{\textbf{0.840} / \textcolor[HTML]{3661AB}{\textbf{0.218}}}{$\pm$ 0.00 / $\pm$ 0.00}{64748B} \\
Cross-modal synthesis & \cellmetric{\underline{0.839} / \textcolor[HTML]{3661AB}{\underline{0.195}}}{$\pm$ 0.00 / $\pm$ 0.00}{64748B} & \cellmetric{0.807 / \textcolor[HTML]{3661AB}{0.144}}{$\pm$ 0.00 / $\pm$ 0.00}{64748B} & \cellmetric{0.782 / \textcolor[HTML]{3661AB}{0.109}}{$\pm$ 0.00 / $\pm$ 0.00}{64748B} & \cellmetric{0.815 / \textcolor[HTML]{3661AB}{0.153}}{$\pm$ 0.00 / $\pm$ 0.00}{64748B} & \cellcolor[HTML]{EBEEFC}\cellmetric{\textbf{0.850} / \textcolor[HTML]{3661AB}{\textbf{0.241}}}{$\pm$ 0.00 / $\pm$ 0.00}{64748B} \\
Pathology description & \cellmetric{\underline{0.861} / \textcolor[HTML]{3661AB}{\underline{0.266}}}{$\pm$ 0.00 / $\pm$ 0.00}{64748B} & \cellmetric{0.819 / \textcolor[HTML]{3661AB}{0.165}}{$\pm$ 0.00 / $\pm$ 0.00}{64748B} & \cellmetric{0.788 / \textcolor[HTML]{3661AB}{0.106}}{$\pm$ 0.00 / $\pm$ 0.00}{64748B} & \cellmetric{0.821 / \textcolor[HTML]{3661AB}{0.164}}{$\pm$ 0.00 / $\pm$ 0.00}{64748B} & \cellcolor[HTML]{EBEEFC}\cellmetric{\textbf{0.863} / \textcolor[HTML]{3661AB}{\textbf{0.314}}}{$\pm$ 0.00 / $\pm$ 0.00}{64748B} \\
Radiology description & \cellmetric{\underline{0.853} / \textcolor[HTML]{3661AB}{\underline{0.239}}}{$\pm$ 0.00 / $\pm$ 0.00}{64748B} & \cellmetric{0.828 / \textcolor[HTML]{3661AB}{0.212}}{$\pm$ 0.00 / $\pm$ 0.00}{64748B} & \cellmetric{0.820 / \textcolor[HTML]{3661AB}{0.164}}{$\pm$ 0.00 / $\pm$ 0.00}{64748B} & \cellmetric{0.821 / \textcolor[HTML]{3661AB}{0.169}}{$\pm$ 0.00 / $\pm$ 0.00}{64748B} & \cellcolor[HTML]{EBEEFC}\cellmetric{\textbf{0.857} / \textcolor[HTML]{3661AB}{\textbf{0.289}}}{$\pm$ 0.00 / $\pm$ 0.01}{64748B} \\
\textbf{Mean} & \cellmetric{\underline{0.841} / \textcolor[HTML]{3661AB}{\underline{0.204}}}{$\pm$ 0.00 / $\pm$ 0.00}{64748B} & \cellmetric{0.808 / \textcolor[HTML]{3661AB}{0.147}}{$\pm$ 0.00 / $\pm$ 0.00}{64748B} & \cellmetric{0.789 / \textcolor[HTML]{3661AB}{0.110}}{$\pm$ 0.00 / $\pm$ 0.00}{64748B} & \cellmetric{0.813 / \textcolor[HTML]{3661AB}{0.149}}{$\pm$ 0.00 / $\pm$ 0.00}{64748B} & \cellcolor[HTML]{EBEEFC}\cellmetric{\textbf{0.851} / \textcolor[HTML]{3661AB}{\textbf{0.258}}}{$\pm$ 0.00 / $\pm$ 0.00}{64748B} \\
\bottomrule
\end{tabular}

\end{table}
 
\begin{table}[t]
\captionsetup{type=table}
\captionof{table}{VQA benchmark results by cancer type on the all-available modality setting. Each cancer is split into MCQ and open-ended rows. MCQ cells report Accuracy\% and macro-F1; open-ended cells report BERTScore-F1 and ROUGE-L F1. Values are mean and standard deviation over inference repeats when available.}
\label{tab:cancer_result}
\centering
\small
\setlength{\tabcolsep}{5pt}
\renewcommand{\arraystretch}{1.18}
\providecommand{\cellmetric}[3]{}
\renewcommand{\cellmetric}[3]{\makebox[\linewidth][c]{\begin{tabular}[t]{@{}c@{}}#1\\[-0.30em]{\scriptsize\textcolor[HTML]{#3}{#2}}\end{tabular}}}
\providecommand{\bertcell}[1]{#1}
\begin{tabular}{>{\raggedright\arraybackslash}p{\dimexpr\linewidth-1.1cm-1.8cm-1.8cm-1.8cm-1.8cm-1.8cm-14\tabcolsep\relax}>{\raggedright\arraybackslash}p{1.1cm}>{\centering\arraybackslash}p{1.8cm}>{\centering\arraybackslash}p{1.8cm}>{\centering\arraybackslash}p{1.8cm}>{\centering\arraybackslash}p{1.8cm}>{\centering\arraybackslash}p{1.8cm}}
\toprule
\rowcolor[HTML]{F3F5FC}
\multicolumn{2}{l}{\textbf{Name}} & GPT-5.4 & \cellcolor[HTML]{EFF1FC}MedGemma & \cellcolor[HTML]{EFF1FC}Gemma-4 & \cellcolor[HTML]{EFF1FC}OncoVLM & \cellcolor[HTML]{EBEEFC}OncoVLM \\
\rowcolor[HTML]{F3F5FC}
\multicolumn{2}{l}{} & -- & \cellcolor[HTML]{EFF1FC}4B & \cellcolor[HTML]{EFF1FC}E4B & \cellcolor[HTML]{EFF1FC}Qwen-3.5-9B & \cellcolor[HTML]{EBEEFC}Qwen-3.5-9B \\
\rowcolor[HTML]{F3F5FC}
\multicolumn{2}{l}{\textbf{Finetuned}} & \faMinus & \cellcolor[HTML]{EFF1FC}\faMinus & \cellcolor[HTML]{EFF1FC}\faMinus & \cellcolor[HTML]{EFF1FC}\faMinus & \cellcolor[HTML]{EBEEFC}\faCheckCircle \\
\rowcolor[HTML]{F3F5FC}
\multicolumn{2}{l}{\textbf{Use projector}} & \faMinus & \cellcolor[HTML]{EFF1FC}\faMinus & \cellcolor[HTML]{EFF1FC}\faMinus & \cellcolor[HTML]{EFF1FC}\faCheckCircle & \cellcolor[HTML]{EBEEFC}\faCheckCircle \\
\noalign{\vskip 0.12em}
\arrayrulecolor[HTML]{000000}\hline\arrayrulecolor{black}
\noalign{\vskip 0.20em}
\rowcolor[HTML]{F3F5FC}
\textbf{Cancer} & \textbf{Group} & \multicolumn{5}{c}{\textbf{Performance by model}} \\
\noalign{\vskip 0.20em}
\arrayrulecolor[HTML]{000000}\hline\arrayrulecolor{black}
\noalign{\vskip 0.28em}
\multirow[c]{2}{=}{\raisebox{-0.55\baselineskip}{\parbox[c]{\linewidth}{\raggedright\textbf{Breast (BRCA)}}}} & MCQ & \cellmetric{\underline{38.3} / \textcolor[HTML]{3661AB}{\underline{0.345}}}{$\pm$ 1.5 / $\pm$ 0.01}{64748B} & \cellmetric{26.4 / \textcolor[HTML]{3661AB}{0.217}}{$\pm$ 0.4 / $\pm$ 0.00}{64748B} & \cellmetric{24.2 / \textcolor[HTML]{3661AB}{0.278}}{$\pm$ 1.5 / $\pm$ 0.02}{64748B} & \cellmetric{32.2 / \textcolor[HTML]{3661AB}{0.279}}{$\pm$ 1.4 / $\pm$ 0.02}{64748B} & \cellcolor[HTML]{EBEEFC}\cellmetric{\textbf{55.6} / \textcolor[HTML]{3661AB}{\textbf{0.543}}}{$\pm$ 1.4 / $\pm$ 0.02}{64748B} \\
 & Open-ended & \cellmetric{\underline{0.842} / \textcolor[HTML]{3661AB}{\underline{0.205}}}{$\pm$ 0.00 / $\pm$ 0.00}{64748B} & \cellmetric{0.811 / \textcolor[HTML]{3661AB}{0.150}}{$\pm$ 0.00 / $\pm$ 0.00}{64748B} & \cellmetric{0.791 / \textcolor[HTML]{3661AB}{0.111}}{$\pm$ 0.00 / $\pm$ 0.00}{64748B} & \cellmetric{0.816 / \textcolor[HTML]{3661AB}{0.151}}{$\pm$ 0.00 / $\pm$ 0.00}{64748B} & \cellcolor[HTML]{EBEEFC}\cellmetric{\textbf{0.855} / \textcolor[HTML]{3661AB}{\textbf{0.274}}}{$\pm$ 0.00 / $\pm$ 0.00}{64748B} \\
\noalign{\vskip 0.30em}
\arrayrulecolor[HTML]{000000}\hline\arrayrulecolor{black}
\noalign{\vskip 0.30em}
\multirow[c]{2}{=}{\raisebox{-0.55\baselineskip}{\parbox[c]{\linewidth}{\raggedright\textbf{Kidney (KIRC)}}}} & MCQ & \cellmetric{\underline{40.2} / \textcolor[HTML]{3661AB}{\underline{0.389}}}{$\pm$ 2.7 / $\pm$ 0.03}{64748B} & \cellmetric{36.7 / \textcolor[HTML]{3661AB}{0.339}}{$\pm$ 0.8 / $\pm$ 0.01}{64748B} & \cellmetric{26.3 / \textcolor[HTML]{3661AB}{0.270}}{$\pm$ 1.1 / $\pm$ 0.01}{64748B} & \cellmetric{24.4 / \textcolor[HTML]{3661AB}{0.268}}{$\pm$ 0.8 / $\pm$ 0.01}{64748B} & \cellcolor[HTML]{EBEEFC}\cellmetric{\textbf{53.9} / \textcolor[HTML]{3661AB}{\textbf{0.541}}}{$\pm$ 2.7 / $\pm$ 0.02}{64748B} \\
 & Open-ended & \cellmetric{\underline{0.848} / \textcolor[HTML]{3661AB}{\underline{0.228}}}{$\pm$ 0.00 / $\pm$ 0.00}{64748B} & \cellmetric{0.812 / \textcolor[HTML]{3661AB}{0.158}}{$\pm$ 0.00 / $\pm$ 0.00}{64748B} & \cellmetric{0.794 / \textcolor[HTML]{3661AB}{0.115}}{$\pm$ 0.00 / $\pm$ 0.00}{64748B} & \cellmetric{0.804 / \textcolor[HTML]{3661AB}{0.149}}{$\pm$ 0.00 / $\pm$ 0.00}{64748B} & \cellcolor[HTML]{EBEEFC}\cellmetric{\textbf{0.854} / \textcolor[HTML]{3661AB}{\textbf{0.272}}}{$\pm$ 0.00 / $\pm$ 0.00}{64748B} \\
\noalign{\vskip 0.30em}
\arrayrulecolor[HTML]{000000}\hline\arrayrulecolor{black}
\noalign{\vskip 0.30em}
\multirow[c]{2}{=}{\raisebox{-0.55\baselineskip}{\parbox[c]{\linewidth}{\raggedright\textbf{Brain (LGG)}}}} & MCQ & \cellmetric{\underline{49.4} / \textcolor[HTML]{3661AB}{\underline{0.490}}}{$\pm$ 3.1 / $\pm$ 0.04}{64748B} & \cellmetric{32.5 / \textcolor[HTML]{3661AB}{0.325}}{$\pm$ 0.2 / $\pm$ 0.00}{64748B} & \cellmetric{24.5 / \textcolor[HTML]{3661AB}{0.246}}{$\pm$ 1.1 / $\pm$ 0.01}{64748B} & \cellmetric{37.4 / \textcolor[HTML]{3661AB}{0.338}}{$\pm$ 1.6 / $\pm$ 0.03}{64748B} & \cellcolor[HTML]{EBEEFC}\cellmetric{\textbf{55.3} / \textcolor[HTML]{3661AB}{\textbf{0.550}}}{$\pm$ 2.1 / $\pm$ 0.02}{64748B} \\
 & Open-ended & \cellmetric{\underline{0.841} / \textcolor[HTML]{3661AB}{\underline{0.206}}}{$\pm$ 0.00 / $\pm$ 0.00}{64748B} & \cellmetric{0.803 / \textcolor[HTML]{3661AB}{0.142}}{$\pm$ 0.00 / $\pm$ 0.00}{64748B} & \cellmetric{0.789 / \textcolor[HTML]{3661AB}{0.110}}{$\pm$ 0.00 / $\pm$ 0.00}{64748B} & \cellmetric{0.820 / \textcolor[HTML]{3661AB}{0.153}}{$\pm$ 0.00 / $\pm$ 0.01}{64748B} & \cellcolor[HTML]{EBEEFC}\cellmetric{\textbf{0.852} / \textcolor[HTML]{3661AB}{\textbf{0.267}}}{$\pm$ 0.00 / $\pm$ 0.00}{64748B} \\
\noalign{\vskip 0.30em}
\arrayrulecolor[HTML]{000000}\hline\arrayrulecolor{black}
\noalign{\vskip 0.30em}
\multirow[c]{2}{=}{\raisebox{-0.55\baselineskip}{\parbox[c]{\linewidth}{\raggedright\textbf{Thyroid (THCA)}}}} & MCQ & \cellmetric{\underline{42.4} / \textcolor[HTML]{3661AB}{\underline{0.411}}}{$\pm$ 1.6 / $\pm$ 0.02}{64748B} & \cellmetric{33.7 / \textcolor[HTML]{3661AB}{0.320}}{$\pm$ 0.6 / $\pm$ 0.01}{64748B} & \cellmetric{17.1 / \textcolor[HTML]{3661AB}{0.168}}{$\pm$ 1.2 / $\pm$ 0.01}{64748B} & \cellmetric{38.2 / \textcolor[HTML]{3661AB}{0.360}}{$\pm$ 2.4 / $\pm$ 0.02}{64748B} & \cellcolor[HTML]{EBEEFC}\cellmetric{\textbf{50.3} / \textcolor[HTML]{3661AB}{\textbf{0.485}}}{$\pm$ 3.0 / $\pm$ 0.03}{64748B} \\
 & Open-ended & \cellmetric{\underline{0.838} / \textcolor[HTML]{3661AB}{\underline{0.192}}}{$\pm$ 0.00 / $\pm$ 0.00}{64748B} & \cellmetric{0.804 / \textcolor[HTML]{3661AB}{0.145}}{$\pm$ 0.00 / $\pm$ 0.00}{64748B} & \cellmetric{0.784 / \textcolor[HTML]{3661AB}{0.111}}{$\pm$ 0.00 / $\pm$ 0.00}{64748B} & \cellmetric{0.820 / \textcolor[HTML]{3661AB}{0.165}}{$\pm$ 0.00 / $\pm$ 0.01}{64748B} & \cellcolor[HTML]{EBEEFC}\cellmetric{\textbf{0.849} / \textcolor[HTML]{3661AB}{\textbf{0.244}}}{$\pm$ 0.00 / $\pm$ 0.00}{64748B} \\
\noalign{\vskip 0.30em}
\arrayrulecolor[HTML]{000000}\hline\arrayrulecolor{black}
\noalign{\vskip 0.30em}
\end{tabular}

\end{table}

\begin{table}[t]
\centering

\begin{minipage}[t]{0.72\linewidth}
\centering
\captionsetup{type=table}
\captionof{table}{Open-ended examples comparing GPT-5.4 and fine-tuned OncoVLM. Red spans mark incorrect or weak GPT evidence, green spans mark OncoVLM evidence aligned with the case, and underlined spans mark the corresponding ground-truth evidence.}
\label{tab:qualitative_open_ended}
\small

\begingroup
\setlength{\tabcolsep}{4.5pt}
\renewcommand{\arraystretch}{1.08}
\pretolerance=0
\tolerance=200
\hyphenpenalty=0
\exhyphenpenalty=0
\doublehyphendemerits=0
\finalhyphendemerits=0
\spaceskip=0.33em plus 0.04em minus 0.04em
\xspaceskip=0.45em plus 0.05em minus 0.04em
\emergencystretch=0pt

\makeatletter
\@ifundefined{hl}{
  \providecommand{\vqahl}[2]{\begingroup\setlength{\fboxsep}{1pt}\colorbox[HTML]{#1}{\strut #2}\endgroup{}}
}{
  \providecommand{\vqahl}[2]{\begingroup\definecolor{vqahlcolor}{HTML}{#1}\sethlcolor{vqahlcolor}\hl{#2}\endgroup{}}
}
\@ifundefined{ul}{
  \providecommand{\vqagtul}[2]{\textcolor[HTML]{#1}{#2}}
}{
  \providecommand{\vqagtul}[2]{\begingroup\definecolor{vqagtulcolor}{HTML}{#1}\setulcolor{vqagtulcolor}\setul{0.42ex}{0.12em}\ul{#2}\endgroup{}}
}
\makeatother

\begin{tabularx}{\linewidth}{
  >{\arraybackslash\footnotesize}X
  >{\arraybackslash\footnotesize}X
  >{\arraybackslash\footnotesize}X
}
\toprule
\rowcolor[HTML]{F3F5FC}
{\normalsize\textbf{GPT-5.4}} &
{\normalsize\textbf{OncoVLM}} &
{\normalsize\textbf{Ground truth}} \\
\midrule

\rowcolor[HTML]{F3F5FC}
\multicolumn{3}{>{\raggedright\arraybackslash\normalsize}p{\dimexpr\linewidth-2\tabcolsep\relax}}{\textbf{Question:} Synthesize pathology and genomics.} \\
...\vqahl{FEE2E2}{spindle-cell} neoplasm with a whorled to storiform architecture, \vqahl{FEE2E2}{bland }\vqahl{FEE2E2}{cytology}, and conspicuous collagenized stroma, without obvious...
&
...aggressive \vqahl{B9EDDD}{diffuse }\vqahl{B9EDDD}{glial} malignancy, showing \vqahl{B9EDDD}{necrosis}, \vqahl{B9EDDD}{microvascular }\vqahl{B9EDDD}{proliferation}, and marked cellular atypia in a high-grade astrocytic...
&
...high-grade \vqagtul{60CAA9}{diffuse }\vqagtul{60CAA9}{glioma} with \vqagtul{60CAA9}{necrosis} and \vqagtul{60CAA9}{microvascular }\vqagtul{60CAA9}{proliferation}, a morphologic pattern concordant with glioblastoma. The genomic... \\

\addlinespace[0.35em]

\rowcolor[HTML]{F3F5FC}
\multicolumn{3}{>{\raggedright\arraybackslash\normalsize}p{\dimexpr\linewidth-2\tabcolsep\relax}}{\textbf{Question:} Describe dominant histology.} \\
...overall low-power appearance is that of a high-grade \vqahl{FEE2E2}{poorly }\vqahl{FEE2E2}{differentiated} urothelial carcinoma, with relatively \vqahl{FEE2E2}{scant} ...
&
...\vqahl{B9EDDD}{papillary }\vqahl{B9EDDD}{urothelial }\vqahl{B9EDDD}{neoplasm} with complex \vqahl{B9EDDD}{branching }\vqahl{B9EDDD}{fronds}, \vqahl{B9EDDD}{fibrovascular }\vqahl{B9EDDD}{cores}, and crowded epithelial proliferation...
&
...\vqagtul{60CAA9}{papillary }\vqagtul{60CAA9}{to }\vqagtul{60CAA9}{nested }\vqagtul{60CAA9}{urothelial }\vqagtul{60CAA9}{neoplasm} with broad fused \vqagtul{60CAA9}{epithelial }\vqagtul{60CAA9}{fronds}, complex [...]  hyalinized \vqagtul{60CAA9}{fibrovascular }\vqagtul{60CAA9}{stroma}... \\

\bottomrule
\end{tabularx}
\endgroup

\end{minipage}
\hfill
\begin{minipage}[t]{0.26\linewidth}
\centering
\captionsetup{type=table}
\captionof{table}{LLM-as-judge on BRCA open-ended. Win rate is the OncoVLM preference rate over GPT-5.4, ties counted as half win.}
\label{tab:llm_as_judge}
\small
\setlength{\tabcolsep}{3.2pt}
\renewcommand{\arraystretch}{1.16}
\newcommand{\judgerowstrut}{\rule[-1.18ex]{0pt}{2.9ex}}

\begin{tabular}{
  >{\raggedright\arraybackslash}p{2.05cm}
  >{\centering\arraybackslash}p{1.05cm}
}
\toprule
\rowcolor[HTML]{F3F5FC}
\textbf{Open-ended task} &
\cellcolor[HTML]{EBEEFC}\textbf{Win rate} \\
\arrayrulecolor[HTML]{000000}\hline\arrayrulecolor{black}
\noalign{\vskip 0.22em}

\judgerowstrut Radiology description &
\cellcolor[HTML]{EBEEFC}\textbf{0.714} \\

\judgerowstrut Pathology description &
\cellcolor[HTML]{EBEEFC}\textbf{0.554} \\

\judgerowstrut Cross-modal synthesis &
\cellcolor[HTML]{EBEEFC}\textbf{0.736} \\

\judgerowstrut Concordance + uncertainty &
\cellcolor[HTML]{EBEEFC}\textbf{0.805} \\

\arrayrulecolor[HTML]{000000}\hline\arrayrulecolor{black}
\textbf{Overall} &
\cellcolor[HTML]{EBEEFC}\textbf{0.702} \\
\bottomrule
\end{tabular}

\end{minipage}

\end{table}

\subsection{Main Results}
\label{sec:main_results}

\textbf{Instruction tuning improves structured clinical prediction.}
Table~\ref{tab:main_result} shows that fine-tuned OncoVLM achieves a 21.8-point accuracy gain over the projector-only base model on Universal MCQs. It also consistently outperforms open medical VLM baselines and surpasses the GPT-5.4 reference on the average Universal MCQ score. This pattern is expected because Universal MCQs are generated from curated clinical and molecular labels with a fixed answer schema. The largest gains occur on structured and joint-profile tasks, suggesting that instruction tuning helps the model map patient-level multimodal evidence to standardized oncology labels.

\textbf{Case-specific MCQs require flexible evidence matching.}
Case-specific MCQs show a different pattern from Universal MCQs. Although fine-tuned OncoVLM improves over the projector-only model, GPT-5.4 remains strongest overall, and open-weight VLMs are closer to OncoVLM than in the universal tasks. This suggests that the difficulty is not simply a failure to learn fixed oncology labels, but the low-structure nature of the case-specific MCQ task itself. Each option is a contrastive fingerprint extracted by comparing the target case with same-cohort cases, so the discriminative cue can vary across examples. Such heterogeneity offers fewer reusable templates for SFT, making task-level adaptation difficult~\citep{chu2025sftmemorizesrlgeneralizes, gudibande2024the}. 

\textbf{LLM-as-judge reveals stronger evidence grounding in open-ended QA.}
For open-ended QA, BERTScore-F1 is tightly clustered across models, while ROUGE-L and LLM-as-judge results show clearer gains for OncoVLM. As shown in Table~\ref{tab:llm_as_judge}, the LLM judge prefers OncoVLM over GPT-5.4 on BRCA open-ended questions with an overall win rate of 0.702, with the strongest gains on concordance and uncertainty, cross-modal synthesis, and radiology description. Further analysis revealed that the judge often favors OncoVLM because GPT-5.4 introduces findings absent from the case summary, whereas OncoVLM stays closer to the available patient evidence. This is consistent with Table~\ref{tab:qualitative_open_ended}, where OncoVLM better preserves discriminative case-specific findings.

\textbf{Cohort-level gains vary with supervision scale and heterogeneity.}
Table~\ref{tab:cancer_result} shows that fine-tuned OncoVLM improves MCQ performance across the reported cancer groups, but the gains are not uniform. Larger cohorts show clearer improvements, while smaller cohorts show smaller margins over GPT-5.4 and sometimes higher variation. This indicates that cohort-level performance is shaped not only by modality availability, but also by the amount and consistency of paired supervision. For example, Brain cancer remains competitive rather than clearly dominated, suggesting that additional cohort-specific supervision may be as important as adding more modalities.

\textbf{Modality ablations show complementary evidence.}
Table~\ref{tab:modality_ablation} shows that different modalities support different task types. The all-available setting performs best on integrative MCQ categories, supporting the value of combining radiology, pathology, and molecular evidence. Pathology-only inputs preserve much of the performance on grade and pathology-description tasks, while radiology-only inputs are more competitive on radiology findings and remain useful for integrated findings. In contrast, open-ended BERTScore-F1 changes only slightly when modalities are removed, again suggesting that surface-level semantic metrics are less sensitive to missing-evidence errors than structured decisions.

\begin{table}[t]

\caption{VQA modality ablation for OncoVLM on cases with pathology and radiology features. MCQ cells report Accuracy\% and macro-F1; open-ended cells report BERTScore-F1 and ROUGE-L F1.}
\label{tab:modality_ablation}
\centering
\small
\setlength{\tabcolsep}{2.5pt}
\renewcommand{\arraystretch}{1.14}
\providecommand{\cellmetric}[3]{}
\renewcommand{\cellmetric}[3]{\makebox[\linewidth][c]{\begin{tabular}[t]{@{}c@{}}#1\\[-0.30em]\textcolor[HTML]{#3}{#2}\end{tabular}}}
\providecommand{\bertcell}[1]{#1}
\begin{tabular}{>{\raggedright\arraybackslash}m{\dimexpr\linewidth-1.32cm-1.32cm-1.32cm-1.32cm-1.32cm-1.32cm-1.32cm-1.32cm-18\tabcolsep\relax}>{\centering\arraybackslash}m{1.32cm}>{\centering\arraybackslash}m{1.32cm}>{\centering\arraybackslash}m{1.32cm}>{\centering\arraybackslash}m{1.32cm}>{\centering\arraybackslash}m{1.32cm}>{\centering\arraybackslash}m{1.32cm}>{\centering\arraybackslash}m{1.32cm}>{\centering\arraybackslash}m{1.32cm}}
\toprule
\rowcolor[HTML]{F3F5FC}
\multicolumn{1}{c}{} & \multicolumn{2}{c}{\begin{tabular}[c]{@{}c@{}}\rule[-0.55ex]{0pt}{3.0ex}\textbf{Universal MCQ}\\[-0.18em]{\small Accuracy\% / \textcolor[HTML]{3661AB}{F1}}\end{tabular}} & \multicolumn{3}{c}{\begin{tabular}[c]{@{}c@{}}\rule[-0.55ex]{0pt}{3.0ex}\textbf{Case-specific MCQ}\\[-0.18em]{\small Accuracy\% / \textcolor[HTML]{3661AB}{F1}}\end{tabular}} & \multicolumn{3}{c}{\begin{tabular}[c]{@{}c@{}}\rule[-0.55ex]{0pt}{3.0ex}\textbf{Case specific open-ended}\\[-0.18em]{\small BERT-F1 / \textcolor[HTML]{3661AB}{ROUGE-L}}\end{tabular}} \\
\arrayrulecolor[HTML]{000000}\hline\arrayrulecolor{black}
\rowcolor[HTML]{F3F5FC}
\textbf{Input} & {\small\textbf{Mutation}} & {\small\textbf{Grade}} & {\small\textbf{Path. findings}} & {\small\textbf{Rad. findings}} & {\small\textbf{Integrated findings}} & {\small\textbf{Path. desc.}} & {\small\textbf{Rad. desc.}} & {\small\textbf{X-modal synth.}} \\
\arrayrulecolor[HTML]{000000}\hline\arrayrulecolor{black}
\noalign{\vskip 0.28em}
\textbf{All} & \cellmetric{\textbf{66.2}}{\textbf{0.660}}{3661AB} & \cellmetric{\textbf{64.0}}{\underline{0.569}}{3661AB} & \cellmetric{\textbf{47.1}}{\textbf{0.460}}{3661AB} & \cellmetric{\underline{32.0}}{\textbf{0.340}}{3661AB} & \cellmetric{\textbf{47.1}}{\textbf{0.463}}{3661AB} & \cellmetric{\textbf{0.861}}{\textbf{0.303}}{3661AB} & \cellmetric{\textbf{0.858}}{\textbf{0.286}}{3661AB} & \cellmetric{\textbf{0.851}}{\textbf{0.257}}{3661AB} \\
\noalign{\vskip 0.16em}
\arrayrulecolor[HTML]{000000}\hline\arrayrulecolor{black}
\noalign{\vskip 0.16em}
\textbf{Path. only} & \cellmetric{50.7}{0.505}{3661AB} & \cellmetric{\textbf{64.0}}{\textbf{0.617}}{3661AB} & \cellmetric{\underline{39.2}}{\underline{0.371}}{3661AB} & \cellmetric{24.0}{0.237}{3661AB} & \cellmetric{37.3}{0.366}{3661AB} & \cellmetric{\underline{0.860}}{\underline{0.295}}{3661AB} & \cellmetric{\underline{0.855}}{\underline{0.283}}{3661AB} & \cellmetric{\underline{0.848}}{\underline{0.242}}{3661AB} \\
\noalign{\vskip 0.16em}
\arrayrulecolor[HTML]{000000}\hline\arrayrulecolor{black}
\noalign{\vskip 0.16em}
\textbf{Rad. only} & \cellmetric{\underline{55.9}}{\underline{0.559}}{3661AB} & \cellmetric{\underline{43.2}}{0.231}{3661AB} & \cellmetric{29.4}{0.285}{3661AB} & \cellmetric{\textbf{32.9}}{\underline{0.328}}{3661AB} & \cellmetric{\underline{42.5}}{\underline{0.424}}{3661AB} & \cellmetric{0.848}{0.259}{3661AB} & \cellmetric{0.854}{0.275}{3661AB} & \cellmetric{0.842}{0.224}{3661AB} \\
\bottomrule
\end{tabular}

\end{table}

\section{Limitations}
\label{sec:limitations}

Interpretation of our results is constrained by the public retrospective cohorts underlying OncoTriad-QA. These cohorts skew toward surgically resectable, treatment-naïve disease and incompletely represent racial and ethnic minorities, older adults, and patients from low- and middle-income countries. Inconsistent clinical annotation and uneven modality coverage further reduce per-cancer statistical power and leave room for cohort-level confounding. Radiology introduces additional heterogeneity through variation in scanner protocols, contrast use, resolution, artifacts, DICOM metadata quality, and MRI sequence labels. Because the benchmark uses selected 2D slices rather than harmonized 3D volumes, imaging biomarkers should be viewed as descriptive grounding signals rather than IBSI-standardized radiomic measurements.

\section{Conclusion}
\label{sec:conclusion}
We introduce OncoTriad-QA, a patient-level radiology-pathology-genomics benchmark for integrated oncology QA, together with OncoVLM, a reference multimodal VLM. Experiments show that OncoVLM improves over general medical and base multimodal baselines, especially on structured clinical tasks, while ablations show different modalities provide complementary signals. Future work will extend the benchmark with volumetric and longitudinal radiology, stronger cross-modal alignment, additional missing-modality settings, and physician-audited adversarial cases.

\begin{ack}
The authors acknowledge support from the University of Central Florida Institute for Artificial Intelligence, including computational infrastructure and large language model API resources used during dataset construction, experimentation, and evaluation.
\end{ack}

\bibliographystyle{plainnat}
\bibliography{references}

\newpage
\appendix

\section{Detailed results}

\begingroup
\captionsetup{type=table}
\captionof{table}{Comprehensive Universal MCQ results by task and cancer on the all-available modality setting. Each task includes up to the top 10 cancer projects with at least 12 questions; remaining projects are pooled as Other. Cells report Accuracy\% / \textcolor[HTML]{3661AB}{F1}. Values are mean and standard deviation over inference repeats when available.}
\label{tab:appendix-mcq-from-ground-truth}
\begingroup
\setlength{\tabcolsep}{3pt}
\renewcommand{\arraystretch}{1.18}
\providecommand{\cellmetric}[3]{}
\renewcommand{\cellmetric}[3]{\makebox[\linewidth][c]{\begin{tabular}[t]{@{}c@{}}#1\\[-0.30em]{\scriptsize\textcolor[HTML]{#3}{#2}}\end{tabular}}}
\providecommand{\bertcell}[1]{#1}
\begin{longtable}{>{\raggedright\arraybackslash}m{\dimexpr\linewidth-1.9cm-1.9cm-1.9cm-1.9cm-1.9cm-12\tabcolsep\relax}>{\centering\arraybackslash}m{1.9cm}>{\centering\arraybackslash}m{1.9cm}>{\centering\arraybackslash}m{1.9cm}>{\centering\arraybackslash}m{1.9cm}>{\centering\arraybackslash}m{1.9cm}}
\toprule
\rowcolor[HTML]{F3F5FC}
\textbf{Name} & GPT-5.4 & \cellcolor[HTML]{EFF1FC}MedGemma & \cellcolor[HTML]{EFF1FC}Gemma-4 & \cellcolor[HTML]{EFF1FC}OncoVLM & \cellcolor[HTML]{EBEEFC}OncoVLM \\
\rowcolor[HTML]{F3F5FC}
 & -- & \cellcolor[HTML]{EFF1FC}4B & \cellcolor[HTML]{EFF1FC}E4B & \cellcolor[HTML]{EFF1FC}Qwen-3.5-9B & \cellcolor[HTML]{EBEEFC}Qwen-3.5-9B \\
\rowcolor[HTML]{F3F5FC}
\textbf{Finetuned} & \faMinus & \cellcolor[HTML]{EFF1FC}\faMinus & \cellcolor[HTML]{EFF1FC}\faMinus & \cellcolor[HTML]{EFF1FC}\faMinus & \cellcolor[HTML]{EBEEFC}\faCheckCircle \\
\rowcolor[HTML]{F3F5FC}
\textbf{Use projector} & \faMinus & \cellcolor[HTML]{EFF1FC}\faMinus & \cellcolor[HTML]{EFF1FC}\faMinus & \cellcolor[HTML]{EFF1FC}\faCheckCircle & \cellcolor[HTML]{EBEEFC}\faCheckCircle \\
\noalign{\vskip 0.12em}
\arrayrulecolor[HTML]{000000}\hline\arrayrulecolor{black}
\noalign{\vskip 0.20em}
\rowcolor[HTML]{F3F5FC}
\textbf{Cancer} & \textbf{GPT-5.4} & \textbf{MedGemma} & \textbf{Gemma-4} & \textbf{OncoVLM} & \textbf{OncoVLM} \\
\midrule
\endfirsthead
\toprule
\rowcolor[HTML]{F3F5FC}
\textbf{Name} & GPT-5.4 & \cellcolor[HTML]{EFF1FC}MedGemma & \cellcolor[HTML]{EFF1FC}Gemma-4 & \cellcolor[HTML]{EFF1FC}OncoVLM & \cellcolor[HTML]{EBEEFC}OncoVLM \\
\rowcolor[HTML]{F3F5FC}
 & -- & \cellcolor[HTML]{EFF1FC}4B & \cellcolor[HTML]{EFF1FC}E4B & \cellcolor[HTML]{EFF1FC}Qwen-3.5-9B & \cellcolor[HTML]{EBEEFC}Qwen-3.5-9B \\
\rowcolor[HTML]{F3F5FC}
\textbf{Finetuned} & \faMinus & \cellcolor[HTML]{EFF1FC}\faMinus & \cellcolor[HTML]{EFF1FC}\faMinus & \cellcolor[HTML]{EFF1FC}\faMinus & \cellcolor[HTML]{EBEEFC}\faCheckCircle \\
\rowcolor[HTML]{F3F5FC}
\textbf{Use projector} & \faMinus & \cellcolor[HTML]{EFF1FC}\faMinus & \cellcolor[HTML]{EFF1FC}\faMinus & \cellcolor[HTML]{EFF1FC}\faCheckCircle & \cellcolor[HTML]{EBEEFC}\faCheckCircle \\
\noalign{\vskip 0.12em}
\arrayrulecolor[HTML]{000000}\hline\arrayrulecolor{black}
\noalign{\vskip 0.20em}
\rowcolor[HTML]{F3F5FC}
\textbf{Cancer} & \textbf{GPT-5.4} & \textbf{MedGemma} & \textbf{Gemma-4} & \textbf{OncoVLM} & \textbf{OncoVLM} \\
\midrule
\endhead
\bottomrule
\endfoot
\arrayrulecolor[HTML]{000000}\hline\arrayrulecolor{black}
\rowcolor[HTML]{F3F5FC}
\multicolumn{6}{l}{\rule[-0.8ex]{0pt}{3.2ex}\textbf{Mutation}} \\
\arrayrulecolor[HTML]{000000}\hline\arrayrulecolor{black}
\noalign{\vskip 0.24em}
Breast (BRCA) & \cellmetric{\underline{46.2} / \textcolor[HTML]{3661AB}{\underline{0.390}}}{$\pm$ 3.4 / $\pm$ 0.03}{64748B} & \cellmetric{\underline{46.2} / \textcolor[HTML]{3661AB}{0.316}}{$\pm$ 0.0 / $\pm$ 0.00}{64748B} & \cellmetric{13.7 / \textcolor[HTML]{3661AB}{0.209}}{$\pm$ 2.0 / $\pm$ 0.03}{64748B} & \cellmetric{45.7 / \textcolor[HTML]{3661AB}{0.367}}{$\pm$ 0.7 / $\pm$ 0.01}{64748B} & \cellcolor[HTML]{EBEEFC}\cellmetric{\textbf{62.4} / \textcolor[HTML]{3661AB}{\textbf{0.621}}}{$\pm$ 2.0 / $\pm$ 0.02}{64748B} \\
Kidney (KIRC) & \cellmetric{52.7 / \textcolor[HTML]{3661AB}{\underline{0.517}}}{$\pm$ 2.5 / $\pm$ 0.03}{64748B} & \cellmetric{\textbf{58.1} / \textcolor[HTML]{3661AB}{0.367}}{$\pm$ 0.0 / $\pm$ 0.00}{64748B} & \cellmetric{26.3 / \textcolor[HTML]{3661AB}{0.277}}{$\pm$ 1.9 / $\pm$ 0.03}{64748B} & \cellmetric{41.4 / \textcolor[HTML]{3661AB}{0.332}}{$\pm$ 0.9 / $\pm$ 0.01}{64748B} & \cellcolor[HTML]{EBEEFC}\cellmetric{\underline{57.0} / \textcolor[HTML]{3661AB}{\textbf{0.566}}}{$\pm$ 7.6 / $\pm$ 0.08}{64748B} \\
Brain (LGG) & \cellmetric{\underline{54.6} / \textcolor[HTML]{3661AB}{\underline{0.527}}}{$\pm$ 1.8 / $\pm$ 0.02}{64748B} & \cellmetric{40.2 / \textcolor[HTML]{3661AB}{0.287}}{$\pm$ 0.0 / $\pm$ 0.00}{64748B} & \cellmetric{22.6 / \textcolor[HTML]{3661AB}{0.322}}{$\pm$ 1.6 / $\pm$ 0.02}{64748B} & \cellmetric{40.9 / \textcolor[HTML]{3661AB}{0.291}}{$\pm$ 0.0 / $\pm$ 0.00}{64748B} & \cellcolor[HTML]{EBEEFC}\cellmetric{\textbf{68.2} / \textcolor[HTML]{3661AB}{\textbf{0.673}}}{$\pm$ 2.8 / $\pm$ 0.03}{64748B} \\
\end{longtable}
\endgroup

\addtocounter{table}{-1}
\endgroup

\begingroup
\captionsetup{type=table}
\captionof{table}{Comprehensive Case-specific MCQ results by task and cancer on the all-available modality setting. Each task includes up to the top 10 cancer projects with at least 12 questions; remaining projects are pooled as Other. Cells report Accuracy\% / \textcolor[HTML]{3661AB}{F1}. Values are mean and standard deviation over inference repeats when available.}
\label{tab:appendix-mcq-from-caption}
\begingroup
\setlength{\tabcolsep}{3pt}
\renewcommand{\arraystretch}{1.18}
\providecommand{\cellmetric}[3]{}
\renewcommand{\cellmetric}[3]{\makebox[\linewidth][c]{\begin{tabular}[t]{@{}c@{}}#1\\[-0.30em]{\scriptsize\textcolor[HTML]{#3}{#2}}\end{tabular}}}
\providecommand{\bertcell}[1]{#1}
\begin{longtable}{>{\raggedright\arraybackslash}m{\dimexpr\linewidth-1.9cm-1.9cm-1.9cm-1.9cm-1.9cm-12\tabcolsep\relax}>{\centering\arraybackslash}m{1.9cm}>{\centering\arraybackslash}m{1.9cm}>{\centering\arraybackslash}m{1.9cm}>{\centering\arraybackslash}m{1.9cm}>{\centering\arraybackslash}m{1.9cm}}
\toprule
\rowcolor[HTML]{F3F5FC}
\textbf{Name} & GPT-5.4 & \cellcolor[HTML]{EFF1FC}MedGemma & \cellcolor[HTML]{EFF1FC}Gemma-4 & \cellcolor[HTML]{EFF1FC}OncoVLM & \cellcolor[HTML]{EBEEFC}OncoVLM \\
\rowcolor[HTML]{F3F5FC}
 & -- & \cellcolor[HTML]{EFF1FC}4B & \cellcolor[HTML]{EFF1FC}E4B & \cellcolor[HTML]{EFF1FC}Qwen-3.5-9B & \cellcolor[HTML]{EBEEFC}Qwen-3.5-9B \\
\rowcolor[HTML]{F3F5FC}
\textbf{Finetuned} & \faMinus & \cellcolor[HTML]{EFF1FC}\faMinus & \cellcolor[HTML]{EFF1FC}\faMinus & \cellcolor[HTML]{EFF1FC}\faMinus & \cellcolor[HTML]{EBEEFC}\faCheckCircle \\
\rowcolor[HTML]{F3F5FC}
\textbf{Use projector} & \faMinus & \cellcolor[HTML]{EFF1FC}\faMinus & \cellcolor[HTML]{EFF1FC}\faMinus & \cellcolor[HTML]{EFF1FC}\faCheckCircle & \cellcolor[HTML]{EBEEFC}\faCheckCircle \\
\noalign{\vskip 0.12em}
\arrayrulecolor[HTML]{000000}\hline\arrayrulecolor{black}
\noalign{\vskip 0.20em}
\rowcolor[HTML]{F3F5FC}
\textbf{Cancer} & \textbf{GPT-5.4} & \textbf{MedGemma} & \textbf{Gemma-4} & \textbf{OncoVLM} & \textbf{OncoVLM} \\
\midrule
\endfirsthead
\toprule
\rowcolor[HTML]{F3F5FC}
\textbf{Name} & GPT-5.4 & \cellcolor[HTML]{EFF1FC}MedGemma & \cellcolor[HTML]{EFF1FC}Gemma-4 & \cellcolor[HTML]{EFF1FC}OncoVLM & \cellcolor[HTML]{EBEEFC}OncoVLM \\
\rowcolor[HTML]{F3F5FC}
 & -- & \cellcolor[HTML]{EFF1FC}4B & \cellcolor[HTML]{EFF1FC}E4B & \cellcolor[HTML]{EFF1FC}Qwen-3.5-9B & \cellcolor[HTML]{EBEEFC}Qwen-3.5-9B \\
\rowcolor[HTML]{F3F5FC}
\textbf{Finetuned} & \faMinus & \cellcolor[HTML]{EFF1FC}\faMinus & \cellcolor[HTML]{EFF1FC}\faMinus & \cellcolor[HTML]{EFF1FC}\faMinus & \cellcolor[HTML]{EBEEFC}\faCheckCircle \\
\rowcolor[HTML]{F3F5FC}
\textbf{Use projector} & \faMinus & \cellcolor[HTML]{EFF1FC}\faMinus & \cellcolor[HTML]{EFF1FC}\faMinus & \cellcolor[HTML]{EFF1FC}\faCheckCircle & \cellcolor[HTML]{EBEEFC}\faCheckCircle \\
\noalign{\vskip 0.12em}
\arrayrulecolor[HTML]{000000}\hline\arrayrulecolor{black}
\noalign{\vskip 0.20em}
\rowcolor[HTML]{F3F5FC}
\textbf{Cancer} & \textbf{GPT-5.4} & \textbf{MedGemma} & \textbf{Gemma-4} & \textbf{OncoVLM} & \textbf{OncoVLM} \\
\midrule
\endhead
\bottomrule
\endfoot
\arrayrulecolor[HTML]{000000}\hline\arrayrulecolor{black}
\rowcolor[HTML]{F3F5FC}
\multicolumn{6}{l}{\rule[-0.8ex]{0pt}{3.2ex}\textbf{Radiology findings}} \\
\arrayrulecolor[HTML]{000000}\hline\arrayrulecolor{black}
\noalign{\vskip 0.24em}
Breast (BRCA) & \cellmetric{\textbf{42.9} / \textcolor[HTML]{3661AB}{\underline{0.298}}}{$\pm$ 14.3 / $\pm$ 0.15}{64748B} & \cellmetric{\textbf{42.9} / \textcolor[HTML]{3661AB}{\textbf{0.367}}}{$\pm$ 14.3 / $\pm$ 0.12}{64748B} & \cellmetric{\underline{33.3} / \textcolor[HTML]{3661AB}{0.250}}{$\pm$ 8.2 / $\pm$ 0.09}{64748B} & \cellmetric{14.3 / \textcolor[HTML]{3661AB}{0.100}}{$\pm$ 0.0 / $\pm$ 0.00}{64748B} & \cellcolor[HTML]{EBEEFC}\cellmetric{23.8 / \textcolor[HTML]{3661AB}{0.214}}{$\pm$ 8.2 / $\pm$ 0.10}{64748B} \\
Kidney (KIRC) & \cellmetric{37.3 / \textcolor[HTML]{3661AB}{0.338}}{$\pm$ 6.8 / $\pm$ 0.06}{64748B} & \cellmetric{\underline{39.2} / \textcolor[HTML]{3661AB}{\underline{0.371}}}{$\pm$ 3.4 / $\pm$ 0.02}{64748B} & \cellmetric{23.5 / \textcolor[HTML]{3661AB}{0.256}}{$\pm$ 0.0 / $\pm$ 0.00}{64748B} & \cellmetric{7.8 / \textcolor[HTML]{3661AB}{0.103}}{$\pm$ 3.4 / $\pm$ 0.03}{64748B} & \cellcolor[HTML]{EBEEFC}\cellmetric{\textbf{43.1} / \textcolor[HTML]{3661AB}{\textbf{0.392}}}{$\pm$ 3.4 / $\pm$ 0.04}{64748B} \\
Liver (LIHC) & \cellmetric{\textbf{66.7} / \textcolor[HTML]{3661AB}{\textbf{0.681}}}{$\pm$ 0.0 / $\pm$ 0.00}{64748B} & \cellmetric{33.3 / \textcolor[HTML]{3661AB}{0.225}}{$\pm$ 0.0 / $\pm$ 0.00}{64748B} & \cellmetric{22.2 / \textcolor[HTML]{3661AB}{0.171}}{$\pm$ 0.0 / $\pm$ 0.00}{64748B} & \cellmetric{0.0 / \textcolor[HTML]{3661AB}{0.000}}{$\pm$ 0.0 / $\pm$ 0.00}{64748B} & \cellcolor[HTML]{EBEEFC}\cellmetric{\underline{37.0} / \textcolor[HTML]{3661AB}{\underline{0.347}}}{$\pm$ 6.4 / $\pm$ 0.05}{64748B} \\
\end{longtable}
\endgroup

\addtocounter{table}{-1}
\endgroup

\begingroup
\captionsetup{type=table}
\captionof{table}{Comprehensive Case specific open-ended results by task and cancer on the all-available modality setting. Each task includes up to the top 10 cancer projects with at least 12 questions; remaining projects are pooled as Other. Cells report BERTScore-F1 / \textcolor[HTML]{3661AB}{ROUGE-L}. Values are mean and standard deviation over inference repeats when available.}
\label{tab:appendix-qa-from-caption}
\begingroup
\setlength{\tabcolsep}{3pt}
\renewcommand{\arraystretch}{1.18}
\providecommand{\cellmetric}[3]{}
\renewcommand{\cellmetric}[3]{\makebox[\linewidth][c]{\begin{tabular}[t]{@{}c@{}}#1\\[-0.30em]{\scriptsize\textcolor[HTML]{#3}{#2}}\end{tabular}}}
\providecommand{\bertcell}[1]{#1}
\begin{longtable}{>{\raggedright\arraybackslash}m{\dimexpr\linewidth-1.9cm-1.9cm-1.9cm-1.9cm-1.9cm-12\tabcolsep\relax}>{\centering\arraybackslash}m{1.9cm}>{\centering\arraybackslash}m{1.9cm}>{\centering\arraybackslash}m{1.9cm}>{\centering\arraybackslash}m{1.9cm}>{\centering\arraybackslash}m{1.9cm}}
\toprule
\rowcolor[HTML]{F3F5FC}
\textbf{Name} & GPT-5.4 & \cellcolor[HTML]{EFF1FC}MedGemma & \cellcolor[HTML]{EFF1FC}Gemma-4 & \cellcolor[HTML]{EFF1FC}OncoVLM & \cellcolor[HTML]{EBEEFC}OncoVLM \\
\rowcolor[HTML]{F3F5FC}
 & -- & \cellcolor[HTML]{EFF1FC}4B & \cellcolor[HTML]{EFF1FC}E4B & \cellcolor[HTML]{EFF1FC}Qwen-3.5-9B & \cellcolor[HTML]{EBEEFC}Qwen-3.5-9B \\
\rowcolor[HTML]{F3F5FC}
\textbf{Finetuned} & \faMinus & \cellcolor[HTML]{EFF1FC}\faMinus & \cellcolor[HTML]{EFF1FC}\faMinus & \cellcolor[HTML]{EFF1FC}\faMinus & \cellcolor[HTML]{EBEEFC}\faCheckCircle \\
\rowcolor[HTML]{F3F5FC}
\textbf{Use projector} & \faMinus & \cellcolor[HTML]{EFF1FC}\faMinus & \cellcolor[HTML]{EFF1FC}\faMinus & \cellcolor[HTML]{EFF1FC}\faCheckCircle & \cellcolor[HTML]{EBEEFC}\faCheckCircle \\
\noalign{\vskip 0.12em}
\arrayrulecolor[HTML]{000000}\hline\arrayrulecolor{black}
\noalign{\vskip 0.20em}
\rowcolor[HTML]{F3F5FC}
\textbf{Cancer} & \textbf{GPT-5.4} & \textbf{MedGemma} & \textbf{Gemma-4} & \textbf{OncoVLM} & \textbf{OncoVLM} \\
\midrule
\endfirsthead
\toprule
\rowcolor[HTML]{F3F5FC}
\textbf{Name} & GPT-5.4 & \cellcolor[HTML]{EFF1FC}MedGemma & \cellcolor[HTML]{EFF1FC}Gemma-4 & \cellcolor[HTML]{EFF1FC}OncoVLM & \cellcolor[HTML]{EBEEFC}OncoVLM \\
\rowcolor[HTML]{F3F5FC}
 & -- & \cellcolor[HTML]{EFF1FC}4B & \cellcolor[HTML]{EFF1FC}E4B & \cellcolor[HTML]{EFF1FC}Qwen-3.5-9B & \cellcolor[HTML]{EBEEFC}Qwen-3.5-9B \\
\rowcolor[HTML]{F3F5FC}
\textbf{Finetuned} & \faMinus & \cellcolor[HTML]{EFF1FC}\faMinus & \cellcolor[HTML]{EFF1FC}\faMinus & \cellcolor[HTML]{EFF1FC}\faMinus & \cellcolor[HTML]{EBEEFC}\faCheckCircle \\
\rowcolor[HTML]{F3F5FC}
\textbf{Use projector} & \faMinus & \cellcolor[HTML]{EFF1FC}\faMinus & \cellcolor[HTML]{EFF1FC}\faMinus & \cellcolor[HTML]{EFF1FC}\faCheckCircle & \cellcolor[HTML]{EBEEFC}\faCheckCircle \\
\noalign{\vskip 0.12em}
\arrayrulecolor[HTML]{000000}\hline\arrayrulecolor{black}
\noalign{\vskip 0.20em}
\rowcolor[HTML]{F3F5FC}
\textbf{Cancer} & \textbf{GPT-5.4} & \textbf{MedGemma} & \textbf{Gemma-4} & \textbf{OncoVLM} & \textbf{OncoVLM} \\
\midrule
\endhead
\bottomrule
\endfoot
\arrayrulecolor[HTML]{000000}\hline\arrayrulecolor{black}
\rowcolor[HTML]{F3F5FC}
\multicolumn{6}{l}{\rule[-0.8ex]{0pt}{3.2ex}\textbf{Concordance + uncertainty}} \\
\arrayrulecolor[HTML]{000000}\hline\arrayrulecolor{black}
\noalign{\vskip 0.24em}
Breast (BRCA) & \cellmetric{\underline{0.820} / \textcolor[HTML]{3661AB}{\underline{0.146}}}{$\pm$ 0.00 / $\pm$ 0.00}{64748B} & \cellmetric{0.796 / \textcolor[HTML]{3661AB}{0.117}}{$\pm$ 0.00 / $\pm$ 0.00}{64748B} & \cellmetric{0.789 / \textcolor[HTML]{3661AB}{0.095}}{$\pm$ 0.00 / $\pm$ 0.00}{64748B} & \cellmetric{0.802 / \textcolor[HTML]{3661AB}{0.131}}{$\pm$ 0.00 / $\pm$ 0.00}{64748B} & \cellcolor[HTML]{EBEEFC}\cellmetric{\textbf{0.841} / \textcolor[HTML]{3661AB}{\textbf{0.219}}}{$\pm$ 0.00 / $\pm$ 0.01}{64748B} \\
Kidney (KIRC) & \cellmetric{\underline{0.828} / \textcolor[HTML]{3661AB}{\underline{0.160}}}{$\pm$ 0.00 / $\pm$ 0.00}{64748B} & \cellmetric{0.798 / \textcolor[HTML]{3661AB}{0.128}}{$\pm$ 0.00 / $\pm$ 0.00}{64748B} & \cellmetric{0.791 / \textcolor[HTML]{3661AB}{0.101}}{$\pm$ 0.00 / $\pm$ 0.00}{64748B} & \cellmetric{0.793 / \textcolor[HTML]{3661AB}{0.129}}{$\pm$ 0.00 / $\pm$ 0.00}{64748B} & \cellcolor[HTML]{EBEEFC}\cellmetric{\textbf{0.842} / \textcolor[HTML]{3661AB}{\textbf{0.222}}}{$\pm$ 0.00 / $\pm$ 0.00}{64748B} \\
Brain (LGG) & \cellmetric{\underline{0.826} / \textcolor[HTML]{3661AB}{\underline{0.150}}}{$\pm$ 0.00 / $\pm$ 0.00}{64748B} & \cellmetric{0.793 / \textcolor[HTML]{3661AB}{0.118}}{$\pm$ 0.00 / $\pm$ 0.00}{64748B} & \cellmetric{0.790 / \textcolor[HTML]{3661AB}{0.108}}{$\pm$ 0.00 / $\pm$ 0.00}{64748B} & \cellmetric{0.809 / \textcolor[HTML]{3661AB}{0.120}}{$\pm$ 0.00 / $\pm$ 0.01}{64748B} & \cellcolor[HTML]{EBEEFC}\cellmetric{\textbf{0.838} / \textcolor[HTML]{3661AB}{\textbf{0.215}}}{$\pm$ 0.00 / $\pm$ 0.01}{64748B} \\
\end{longtable}
\endgroup

\addtocounter{table}{-1}
\endgroup

\section{Genomics Text Generation Pipeline}
\label{appendix:genomics-text}

For each patient, raw molecular data from the GDC are converted into a structured genomics text block consumed by the teacher model during case-narrative and QA generation. The pipeline is LLM-generation-free: every field is the deterministic output of a summarization function---panel selection, gene-set scoring, threshold-based discretization, signature averaging, or subtype classification---rendered into a fixed-order template. Missing or insufficient inputs are emitted as explicit \texttt{not\_assessed} markers rather than dropped, preserving the distinction between ``not tested'' and ``tested negative.'' 
 
\textbf{Inputs and per-case selection.}
Five GDC modalities are used: methylation beta tables, gene-level copy-number tables, segment-level copy-number tables, masked somatic mutation calls (MAF), and miRNA quantifications, with bulk RNA-seq and external fusion calls. A unified per-patient registry resolves one file per modality, preferring tumor over normal samples for RNA-seq, masked ensemble MAFs for mutations, and ASCAT- or GISTIC-style outputs for copy number.
 
\textbf{Methylation features.}
After probe-level normalization, we compute promoter methylation by gene over curated cohort-specific tumor-suppressor panels (e.g., \emph{VHL}, \emph{CDKN2A}, \emph{MLH1}, \emph{BRCA1}), discretized into low, intermediate, or high; a CIMP call for cohorts with established CIMP definitions (colorectal, gastric, endometrial, esophageal); a global methylation mean; Horvath epigenetic age and age acceleration when at least 300 of 353 clock CpGs are present; and a LUMP-style methylation-derived tumor purity.
 
\textbf{Transcriptome features.}
From log-transformed STAR-Counts TPMs over Ensembl protein-coding genes, we compute single-sample enrichment for the 50 MSigDB Hallmark sets (top five enriched, bottom three suppressed are emitted), mean expression signatures for proliferation, hypoxia, EMT, IFN-$\gamma$ response, the tumor inflammation signature, and cytolytic activity, ESTIMATE-style stromal/immune/purity scores, marker-panel scores for major immune and stromal cell types, PAM50 subtype calls for breast adenocarcinoma, and recurrent cohort-relevant fusions when fusion calls are available.
 
\textbf{Mutation and copy-number features.}
MAFs are filtered to a per-cohort driver-gene panel and to non-silent, protein-altering variant classes. For each panel gene we emit either a mutation entry with standardized protein change, loss-of-function flag, and OncoKB-style hotspot flag, or an explicit wild-type call. TMB is computed as non-silent variants per 38\,Mb and binned. Gene-level copy number is discretized into GISTIC-compatible deep deletion, shallow loss, neutral, gain, or amplification calls. Segment-level files are mapped to hg38 chromosome arms with coverage-weighted means and filtered to recurrent, prognostically informative cohort-specific arm events (e.g., 3p loss in ccRCC, 1p/19q codeletion in LGG, 5q gain in colorectal). MSI and HRD status are passed through from registry metadata when available.
 
\textbf{Integrated surrogates.}
Cross-stream surrogates are derived from the per-modality feature objects: an MSI-like signal combining \emph{MLH1} hypermethylation with IFN-$\gamma$; an HRD-like signal combining \emph{BRCA1} hypermethylation with \emph{BRCA1}/\emph{BRCA2} mutation; hormone-receptor concordance from \emph{ESR1}, \emph{PGR}, \emph{ERBB2}; \emph{VHL} pathway inactivation combining mutation, promoter methylation, and the hypoxia signature; and a CIMP surrogate carried into the text-channel section. Each surrogate is rendered with its constituent inputs.

\section{Radiology Image Processing Pipeline}
\label{appendix:radiology}
This section describes the full path that radiology data takes in OncoTriad-QA, from raw download to the per-case biomarker text block consumed by the teacher model and used as privileged supervision for OncoVLM. The pipeline has five steps: download, DICOM-level filtering, PNG rendering, tumor-only segmentation with Medical SAM3, and biomarker extraction with series- and case-level aggregation.

\textbf{Download and modality scope.}
For each case across the 32 cohorts, we pull radiology studies from The Cancer Imaging Archive (TCIA) and keep only computed tomography (CT) and magnetic resonance (MR); positron emission tomography, mammography, nuclear medicine, and plain radiography are dropped at this stage. A single case may contain several studies, and each study contains one or more series. Each series is already stored as a stack of 2D DICOM slices, which we treat as the unit of organization throughout the pipeline.

\textbf{DICOM-level filtering.}
Many series in TCGA radiology cohorts are not suitable for tumor-focused analysis: scout views, dose reports, derived secondary captures, segmentation overlays, key-image stacks, and series with too few slices, missing pixel data, or inconsistent geometry. Before any image processing, we read each series' DICOM headers and discard series that fail simple structural checks (modality outside CT/MR, fewer than a minimum slice count, missing required tags, non-image SOP classes, or unreadable pixel arrays). This step removes a large fraction of series and prevents downstream stages from spending compute on unusable data.

\textbf{Slice rendering to PNG.}
The remaining DICOM slices are rendered to PNG so that the teacher model and the radiology encoder see the same input. For CT, the modality lookup-table is applied to recover Hounsfield units, and the slice is rendered into an RGB image whose three channels carry soft-tissue, lung, and bone window settings, preserving attenuation information at three clinically relevant ranges in a single image. For MR, where no absolute calibration exists, intensities are clipped to the 1st--99th percentile and min--max normalized. Background padding is masked before windowing, and slices are resized to the radiology encoder's native input size.

\textbf{Tumor segmentation with Medical SAM3.}
Every rendered PNG is segmented with Medical SAM3~\citep{medicalsam3}, with frozen weights and a single text prompt, \texttt{tumor}, used uniformly across CT, MR, and all 32 cohorts. We do not segment any other structure: no organ masks, no vessel masks, no anatomical priors. The predicted probability map is binarized and small connected components below a fixed pixel-count threshold are removed to suppress noise. Each slice therefore yields a candidate tumor mask and associated quality flags, which is the only segmentation signal used for the rest of the pipeline.

\textbf{Biomarker extraction from tumor masks.}
For each slice, we compute a structured biomarker record from the tumor mask and the underlying rendered image, restricted to the tumor region and a thin band immediately around it. Three groups of features are extracted: (i) \emph{tumor-mask geometry}, including area fraction, equivalent diameter, principal axes, eccentricity, circularity, solidity, extent, perimeter ratios, boundary irregularity, and fragmentation; (ii) \emph{boundary and rim descriptors}, computed from a thin eroded rim band whose width is scaled to the tumor's equivalent radius, capturing the peripheral-to-core intensity ratio and a Sobel-based boundary sharpness score normalized by within-tumor intensity variation; (iii) \emph{texture and intensity statistics inside the tumor}, including Shannon entropy, masked grey-level co-occurrence matrix features (contrast, homogeneity, energy, correlation at four canonical angles), higher-order moments (skewness, kurtosis), and robust spread statistics (median absolute deviation, IQR, RMS). A necrosis proxy is also computed as the fraction of within-tumor pixels in the lowest intensity quartile, together with how centrally that low-intensity region sits inside the mask. For CT, per-channel statistics are reported separately for the soft-tissue, lung, and bone windows; for MR, within-tumor statistics are expressed relative to the same-slice non-tumor distribution. All values are deterministic functions of the tumor mask and the rendered slice; nothing draws on raw Hounsfield values or DICOM headers that are not recoverable from the PNG. Each record also carries a parallel block of segmentation-quality flags (empty, very small, fragmented, or edge-touching mask), which travel alongside the numeric fields rather than gating inclusion.

\textbf{Series- and case-level aggregation.}
Each series is summarized by combining its slice-level records, and a per-series quality score is computed from the fraction of slices with valid masks and the mean tumor area fraction. For every case, we rank all eligible series by this score and keep the top three, consistent with $k{=}3$ in the main paper. Within each retained series, we keep two representative 2D slices: the middle slice of the series and the slice whose tumor mask has the largest area. These are the slices used both for biomarker reduction and as visual inputs to the teacher model. Slice-level records are averaged into a per-series vector; the up to three retained per-series vectors are then combined into a single case-level vector by a weighted mean whose weights are the per-series quality scores, so series with cleaner segmentations and larger tumor visibility contribute proportionally more. Null fields are dropped before weight renormalization, so features computable on only a subset of series remain represented at the case level. The final case-level vector is rendered into a fixed-schema text block consumed by the teacher model during case-narrative generation and, under the privileged-information framing of OncoVLM, as supervised context during student training.

\section{Experimental Setup Details}
\label{app:experimental_setup}

All training, validation, and test partitions used in this work come from a single source-of-truth assignment that labels each row of OncoTriad-QA prior to any downstream stage. Both projector pretraining and instruction tuning consume the same labels, and no training stage ever re-splits data. Splits are deterministic and patient-level: a stable hash of the (source, cohort, patient) identifier is bucketed into training, validation, and test in the ratios $0.85$:$0.05$:$0.10$.

\subsection{Stage 1: modality projector pretraining}
\label{app:stage1}
 
In Stage~1, each modality projector is trained independently while both its modality-specific encoder and the language-model backbone are frozen. The optimization objective is a next-token cross-entropy on the supervision text (modality-specific captions or short case descriptions), conditioned on the projected modality tokens prepended to the prompt. We do not tie projectors across modalities, do not share optimizers, and do not interleave updates: each modality has its own run with its own learning rate, schedule, and effective batch size.
All four projectors target the same Qwen3.5-9B token embedding space so that they can be loaded together in Stage~2.
 
\textbf{Compute}
 We use a single NVIDIA A100 (40 GB) GPU per projector training. with a per-projector wall-clock of approximately 12–18 hours, giving a total Stage 1 cost of roughly 60 hours across the four projectors
 
\textbf{Projector architectures.}
The pathology branch produces a long, variable-length set of patch embeddings per slide and uses a Perceiver-style cross-attention resampler with $256$ learned latents, depth $2$, $8$ heads, and dropout $0.1$, compressing slide-level evidence to a fixed token budget. Radiology, RNA-seq, and DNA methylation projectors are two-layer MLPs with dropout $0.05$ between projection and the embedding space; the DNA-methylation projector additionally prepends $8$ learned prefix tokens.
 
\textbf{Optimizer, schedule, and seed.}
All four projectors are optimized with AdamW with weight decay $0.01$, gradient clipping at $\ell_2$ norm $1.0$, and a cosine learning-rate schedule with linear warmup. Optimizer state is initialized from scratch for every projector run; we do not warm-start from any other modality. The global random seed is $42$, fixed across PyTorch, NumPy, and CUDA generators, and propagated to the dataloader workers.
 
\textbf{Per-modality settings and validation.}
Per-modality settings are summarized in Table~\ref{tab:stage1-hparams}. Radiology,
RNA, and methylation projectors use the benchmark-provided training and validation
splits. The pathology projector trains only on the training split and constructs
an internal slide-level holdout equal to $5\%$ of the training rows for early
stopping; we use this finer-grained holdout because pathology supervision is
slide-level rather than case-level, and matching the validation granularity to the
supervision unit yielded more stable convergence in development. Best checkpoints
are selected by lowest validation loss within the configured epoch budget.
 
\begin{table}[h]
\centering
\small
\caption{Stage~1 projector pretraining hyperparameters. All four projectors target
the Qwen3.5-9B embedding space; the language model and modality encoder are frozen
during projector training. Effective batch size is the product of the per-step
batch size and the gradient accumulation factor.}
\label{tab:stage1-hparams}
\begin{tabular}{lcccc}
\toprule
& Pathology & Radiology & DNAm & RNA \\
\midrule
Architecture           & Resampler & MLP & MLP + 8 prefix & MLP \\
Latents / hidden       & 256       & --  & --             & -- \\
Depth / layers         & 2         & 2   & 2              & 2  \\
Attention heads        & 8         & --  & --             & -- \\
Dropout                & 0.10      & 0.05& 0.05           & 0.05 \\
\midrule
Epochs                 & 16        & 10   & 30   & 20 \\
Batch size             & 1         & 2    & 4    & 4  \\
Grad. accumulation     & 32        & 8    & 8    & 8  \\
Effective batch size   & 32        & 16   & 32   & 32 \\
Optimizer              & AdamW     & AdamW& AdamW& AdamW \\
Learning rate          & $1\!\times\!10^{-4}$ & $5\!\times\!10^{-5}$ & $5\!\times\!10^{-5}$ & $5\!\times\!10^{-5}$ \\
Weight decay           & 0.01      & 0.01 & 0.01 & 0.01 \\
LR schedule            & cosine    & cosine & cosine & cosine \\
Warmup ratio           & 0.05      & --   & --   & -- \\
Grad. clip ($\ell_2$)  & 1.0       & 1.0  & 1.0  & 1.0 \\
Validation             & internal 5\% slide & benchmark val & benchmark val & benchmark val \\
Selection metric       & val loss  & val loss & val loss & val loss \\
\bottomrule
\end{tabular}
\end{table}
 
\subsection{Stage 2: instruction tuning with LoRA}
\label{app:stage2}
 
\textbf{Trainable parameters.}
In Stage~2, the modality encoders, all four pretrained projectors, the LM
embeddings, and the LM head are frozen. The only trainable parameters are LoRA
adapters inserted into the seven attention and MLP projections of every transformer
block of Qwen3.5-9B, with rank $r=8$, scaling $\alpha=32$, dropout $0.05$, and zero
bias. LoRA is applied to the query, key, value, and output projections of
self-attention and to the gate, up, and down projections of the MLP, giving 14,548,992 trainable parameters out of $\sim$9.0B base parameters.
 
\textbf{Compute}
 We use a two NVIDIA A100 (80 GB) GPU for the stage 2 training with the both training time of around 24 hours.
 
\textbf{Optimizer, schedule, and seed.}
We use AdamW with learning rate $1\!\times\!10^{-4}$, weight decay $0$, cosine
schedule, warmup ratio $0.03$, and gradient clipping at $\ell_2$ norm $1.0$. The
per-step batch size is $1$ with gradient accumulation $16$, giving an effective
batch size of $32$. Training runs for $3$ epochs on the training split; we monitor
validation loss but do not early-stop, so the duration of every Stage~2 run is
fixed by the epoch budget. The maximum tokenized text length is $1024$, and the
random seed is $42$, fixed across all stochastic components including dataloader
shuffling.

\textbf{Loss and target masking.}
The loss is the standard autoregressive next-token cross-entropy, computed only
over target answer tokens. Instruction text and modality-prefix tokens are masked
out of the loss but contribute to the conditioning context. For MCQs, the target
is the exact text of the correct option; for open-ended items, the target is the
GPT-5.4 reference answer. We do not apply label smoothing, and we do not use any
reward or preference-based objective.
 
\subsection{Inference and generation protocol}
\label{app:inference}
 
\textbf{Stratified evaluation sample.}
The numbers reported in the main paper are computed on a stratified subsample of the held-out test split, drawn once and frozen before any model is evaluated. OncoTriad-QA expands each held-out case into multiple item variants by enumerating the modality-availability configurations relevant to that case (all-available, pathology-only, radiology-only), and reported number is the mean over three independent inference repeats with different seeds so that we can quote standard deviations. We draw a stratified sample covering $60\%$ of the post-expansion test items and use exactly the same sample for every system, and every inference repeat. Stratification is applied jointly along two axes: (i) cancer cohort, so that per-cohort tables are not dominated by a single project; (ii) question family, so that the distribution over universal MCQs, case-specific MCQs, and case-specific open-ended items matches the full split. Sampling is deterministic, seeded from the same global seed used elsewhere in the pipeline.
 
\textbf{Image-input handling at inference.}
For models that consume raw images (MedGemma-4B, Gemma-4-E4B, GPT-5.4), we cap input at most $8$ pathology PNGs and at most $8$ radiology PNGs per case, choosing the canonical representative slices and tiles produced by the dataset construction pipeline.
 
\subsection{Baseline configurations}
\label{app:baselines}
 
Each baseline is run with the configuration in Table~\ref{tab:baselines}.
 
\begin{table}[h]
\centering
\small
\caption{Baseline operating configurations at inference time. Quantization, dtype,
and attention backends apply only to locally hosted models; GPT-5.4 is accessed
through Azure OpenAI.}
\label{tab:baselines}
\begin{tabular}{lllllc}
\toprule
Model & Backend & Quant. & Dtype & Prompt profile & Batch \\
\midrule
GPT-5.4                 & Azure OpenAI         & --      & --     & baseline       & 8 \\
MedGemma-4B-it          & Hugging Face         & fp16    & bf16   & baseline       & 10 \\
Gemma-4-E4B-it          & Hugging Face         & 8-bit   & bf16   & baseline       & 6  \\
OncoVLM-base (proj.)    & local                & 8-bit   & bf16   & baseline       & 16 \\
OncoVLM-finetuned (LoRA)& local                & 8-bit   & bf16   & OncoVLM        & 16 \\
\bottomrule
\end{tabular}
\end{table}

\section{Licenses for Existing Assets}
\label{app:licenses}

OncoTriad-QA and OncoVLM build on a number of publicly released datasets, pretrained models, and software libraries. We summarize each asset, the version we used, the redistribution license or terms of use that govern it, and the access URL. All assets were used in compliance with their stated licenses; controlled-access tiers were not used.

\subsection{Datasets}
\label{app:licenses:data}

\textbf{The Cancer Genome Atlas (TCGA) program.}
TCGA \citep{tcga} (\url{https://www.cancer.gov/ccg/research/genome-sequencing/tcga}) is the underlying source program for all patient cohorts in OncoTriad-QA. TCGA  remains publicly available for research use through its successor repositories (GDC for molecular and clinical data, TCIA for imaging). All cases used in this work were originally consented under the TCGA informed consent framework for research use only; we did not re-consent or re-contact any participants. Use of TCGA-derived data is governed by the TCGA Publication Guidelines (\url{https://www.cancer.gov/ccg/research/genome-sequencing/tcga/using-tcga/citing-tcga}) and the data-tier-specific policies of the hosting repository (GDC or TCIA, below). TCGA program attribution is required in any downstream publication that uses
TCGA-derived cohorts, including OncoTriad-QA.

\textbf{Genomic Data Commons (GDC) molecular and clinical data.}
Source: NCI Genomic Data Commons \citep{gdc} at \url{https://portal.gdc.cancer.gov}. The GDC hosts the harmonized TCGA molecular and clinical data used in OncoTriad-QA. We use only the \emph{Open Access} tier, which contains de-identified molecular and clinical data that cannot be attributed to an individual research participant. This includes masked somatic mutations, gene-level and segment-level copy-number, DNA methylation beta values, STAR-Counts RNA-seq quantifications, miRNA quantifications, and clinical/biospecimen supplements (BCR XML). No controlled-access genomic data (e.g.~raw BAMs, germline variants, SNP6 genotypes) were used. Use of GDC Open Access data is governed by the NIH Genomic Data Sharing Policy (\url{https://gdc.cancer.gov/access-data/data-access-policies}); attribution to both the GDC \citep{gdc} and the underlying TCGA program \citep{tcga} is required.

\textbf{The Cancer Imaging Archive (TCIA) radiology.}
Source: TCIA \citep{tcia} at \url{https://www.cancerimagingarchive.net}. TCIA hosts the TCGA radiology imaging used in OncoTriad-QA. TCIA uses per-collection licenses (Citation \& Data Usage Policy on each collection's landing page). For the TCGA imaging cohorts used in this paper, the prevailing license is the Creative Commons Attribution (CC~BY) 3.0 Unported or 4.0 International license, which permits commercial, scientific, and educational reuse with attribution to both the individual TCGA collection DOI and the TCIA archive itself \citep{tcia}

\textbf{TCGA-UniformTumor-8K (TCGA-UT-8K).}
We use the publicly released ROI tiles from TCGA-UT-8K, distributed alongside TITAN \citep{titan} at \url{https://huggingface.co/datasets/MahmoodLab/TCGA-UT-8K}. The dataset is released under the CC~BY-NC-ND~4.0 license and may only be used for non-commercial, academic research, with attribution and without distribution of derivative tiles. We use the released tiles only as supplementary visual context for the teacher model during case-narrative generation; we do not redistribute the tiles or any tile-level derivatives.

\subsection{Pretrained Encoders and Segmentation Model}
\label{app:licenses:encoders}

\textbf{MedSigLIP-448 (radiology encoder).}
\texttt{google/medsiglip-448} on Hugging Face (\url{https://huggingface.co/google/medsiglip-448}). The model weights are governed by the Health AI Developer Foundations (HAI-DEF) Terms of Use (\url{https://developers.google.com/health-ai-developer-foundations/terms}) and Prohibited Use Policy (\url{https://developers.google.com/health-ai-developer-foundations/prohibited-use-policy}).
HAI-DEF is an open-weight (not open-source) license that permits research and commercial use subject to the listed restrictions, requires distribution of the HAI-DEF notice with any derivative, and disclaims medical-device authorization. The companion repository code at \url{https://github.com/Google-Health/medsiglip} is licensed under Apache~2.0. We use MedSigLIP-448 as a frozen radiology encoder \citep{medgemma}; no weight modifications are performed.

\textbf{UNI (pathology encoder).}
\texttt{MahmoodLab/UNI} on Hugging Face (\url{https://huggingface.co/MahmoodLab/UNI}; code at \url{https://github.com/mahmoodlab/UNI}) \citep{uni}. The model weights and associated code are released under the CC~BY-NC-ND~4.0 license and are limited to non-commercial, academic research with proper attribution.

\textbf{BulkFormer (transcriptome encoder).}
Source code and pretrained weights at \url{https://github.com/KangBoming/BulkFormer} \citep{bulkformer}, with associated data archived at \url{https://doi.org/10.5281/zenodo.15559368}.
The repository is licensed under the MIT License, which permits unrestricted commercial and non-commercial use with attribution. We use the released \textsc{BulkFormer} checkpoint as a frozen encoder.

\textbf{CpGPT (DNA methylation encoder).}
Source code and pretrained weights at \url{https://github.com/lucascamillomd/CpGPT} \citep{cpgpt}. CpGPT is released under the MIT License.

\textbf{Medical SAM3 (radiology segmentation).}
Code at \url{https://github.com/AIM-Research-Lab/Medical-SAM3} and weights at \url{https://huggingface.co/Chongcong/Medical-SAM3} \citep{medicalsam3}. At the time of submission, the repository does not contain a top-level \texttt{LICENSE} file or an explicit license declaration in its model card; we therefore use the model only for research evaluation and tumor-mask generation, do not redistribute its weights, and do not include any Medical SAM3 outputs that would conflict with any downstream license that is ultimately attached.

\subsection{Language Model Backbones and Baselines}
\label{app:licenses:llms}

\textbf{Qwen3.5-9B (OncoVLM backbone).}
\texttt{Qwen/Qwen3.5-9B} on Hugging Face (\url{https://huggingface.co/Qwen/Qwen3.5-9B}). Released under the Apache~2.0 license, which permits commercial and non-commercial use, modification, and
redistribution with attribution, subject to the standard patent and trademark clauses.

\textbf{MedGemma-4B (baseline).}
\texttt{google/medgemma-4b-it} on Hugging Face
(\url{https://huggingface.co/google/medgemma-4b-it}) \citep{medgemma}. Governed by the HAI-DEF Terms of Use and Prohibited Use Policy (same terms as MedSigLIP-448). Used only for zero-shot evaluation; no weights are redistributed.

\textbf{Gemma-4 4B Effective (Gemma-4 E4B in tables).}
The base \texttt{google/gemma-4-E4B} model is governed by the Gemma Terms of Use (\url{https://ai.google.dev/gemma/terms}), which permit research and commercial use with the listed restrictions and require propagation of the Gemma terms to downstream recipients. Used only for zero-shot evaluation.

\textbf{GPT-5.4 (teacher and zero-shot baseline).}
Accessed via the OpenAI API \citep{gpt54} at \url{https://platform.openai.com}. Use is governed by the OpenAI Business Terms and Usage Policies in effect at the time of access. The teacher model is used to generate case narratives and downstream QA pairs ; we redistribute the resulting \emph{annotations} (text), but not the model itself.

\section{Broader Impact}
\label{app:impact}

OncoTriad-QA and OncoVLM are released as research artifacts to support the development and evaluation of multimodal oncology models. We discuss both the intended positive impacts and the foreseeable risks below.

\textbf{Positive impacts.} A standardized patient-level benchmark that aligns radiology, pathology, and genomics across 32 cancer cohorts can accelerate research on integrated cancer reasoning by giving the community a common ground for comparison, ablation, and missing-modality evaluation. By formulating cancer characterization as evidence-grounded QA rather than narrow modality-specific tasks, OncoTriad-QA encourages models that attend to the same heterogeneous evidence streams used in clinical oncology, and the accompanying reference model demonstrates how frozen modality encoders can be combined with an LLM backbone in a reproducible way. Open release of code, structured case representations, and evaluation splits also lowers the barrier for groups without proprietary multimodal data, including academic and resource-constrained settings.

\textbf{Risks and limitations of impact.} The most direct risk is inappropriate clinical use. OncoTriad-QA is built from retrospective public data that emphasizes surgically removable, treatment-naïve cancers and underrepresents racial and ethnic minorities, elderly patients, and patients from low- and middle-income countries; models tuned on this distribution should not be treated as clinical decision-support without prospective validation. Because case-specific narratives and QA items are produced with teacher-model assistance, residual hallucinations or distributional artifacts may persist despite consistency checks and physician audit, and could be amplified by downstream models trained on these annotations. Surface-level semantic metrics such as BERTScore-F1 may further mask evidence-grounding errors, as our open-ended results show. A secondary concern is dual use: the integrated radiology--pathology--genomics framing could in principle be repurposed to infer sensitive patient attributes from public archives, although the underlying TCIA/TCGA/GDC data are already de-identified and access-controlled at the source.

\textbf{Mitigations.} We release OncoTriad-QA explicitly as a research benchmark rather than a clinical tool, document modality coverage, cohort imbalance, and known distributional limitations (Section \ref{sec:limitations}), and provide stratified splits by cohort and modality availability so that reviewers can audit per-cohort behavior. Annotations carry explicit \texttt{not\_assessed} markers and uncertainty cues to discourage overconfident generation, and a subset of cases were reviewed by a physician to refine the generation guidelines toward evidence-grounded wording. Users of the released artifacts are expected to comply with the original data-use agreements of TCIA, TCGA, and GDC, and to clearly distinguish benchmark performance from clinical performance in any downstream reporting.



\end{document}